\documentclass[11pt]{article}
\usepackage{graphicx}% Include figure files
\usepackage{dcolumn}% Align table columns on decimal point
\usepackage{bm}% bold math
\usepackage{epsfig}
\usepackage{subfigure}
\usepackage{my-apa-uiuc}
\usepackage{moreverb}
\begin{document}
\begin{titlepage}
\vspace*{1.6in}
\begin{center}
\hspace*{1.2cm}{\large{\bf Population Sizing for Genetic Programming }}\\ 
\hspace*{1.2cm}{\large{\bf Based Upon Decision Making }}\\
\hspace*{1.2cm}~\\
\hspace*{1.2cm}~\\
\hspace*{1.2cm}{\bf Kumara Sastry}\\
\hspace*{1.2cm}{\bf Una-May O'Reilly}\\
\hspace*{1.2cm}{\bf David E. Goldberg}\\
\hspace*{1.2cm}~\\
\hspace*{1.2cm}~\\
\hspace*{1.2cm}~\\
\hspace*{1.2cm}IlliGAL Report No. 2004028\\
\hspace*{1.2cm}April, 2004\\
\vspace*{3in}
{\large
\hspace*{1.2cm}Illinois Genetic Algorithms Laboratory (IlliGAL)\\
\hspace*{1.2cm}Department of General Engineering\\
\hspace*{1.2cm}University of Illinois at Urbana-Champaign \\
\hspace*{1.2cm}117 Transportation Building \\
\hspace*{1.2cm}104 S. Mathews Avenue, Urbana, IL 61801 \\}
\end{center}
\end{titlepage}

\title{Population Sizing for Genetic Programming Based Upon Decision Making}
\author{
Kumara Sastry\\
Illinois Genetic Algorithms Laboratory (IlliGAL), and\\
Department of Material Science \& Engineering\\
University of Illinois at Urbana-Champaign\\
{\tt ksastry@uiuc.edu}
\and
Una-May O'Reilly\\
Computer Science \& Artificial Intelligence Laboratory\\
Massachusetts Institute of Technology, Cambridge, MA, USA\\
{\tt unamay@csail.mit.edu}
\and
David E. Goldberg\\
Illinois Genetic Algorithms Laboratory (IlliGAL), and\\
Department of General Engineering\\
University of Illinois at Urbana-Champaign\\
{\tt deg@uiuc.edu}}
\date{}
\maketitle

\begin{abstract}
  This paper derives a population sizing relationship for genetic
  programming (GP). Following the population-sizing derivation for
  genetic algorithms in \citeN{Goldberg:1992:ComplexSystems}, it
  considers building block decision making as a key facet. The
  analysis yields a GP-unique relationship because it has to account
  for bloat and for the fact that GP solutions often use subsolutions
  multiple times. The population-sizing relationship depends upon tree
  size, solution complexity, problem difficulty and building block expression
  probability.  The relationship is used to analyze and empirically
  investigate population sizing for three model GP problems named {\tt
  ORDER}, {\tt ON-OFF} and {\tt LOUD}. These problems exhibit bloat to
  differing extents and differ in whether their solutions require the
  use of a building block multiple times.
\end{abstract}

 \newcommand{\szm}{$\lambda$}
 \newcommand{\sz}{\lambda}
 \newcommand{\trials}{\tau}
 \newcommand{\trialsm}{$\tau$}                                                    

%Let's test this:
%\begin{description}
%\item[ ] This is size with implicit math env: \szm{}. Use  $\sz{}$ inside math environment 
%\item[ ] This is trials with implicit math env: \trialsm{}. Use  $\trials{}$ inside math environment (equation)
%\end{description}

\section{Introduction}
The growth in application of genetic programming (GP) to problems of
practical and scientific importance is remarkable
\cite{keijzer:2004:GP,RioloWorzel:2003,GECCO2003-PartI,GECCO2003-PartII}.
Yet, despite this increasing interest and empirical success, GP
researchers and practitioners are often frustrated---sometimes
stymied---by the lack of theory available to guide them in selecting
key algorithm parameters or to help them explain empirical findings in
a systematic manner. For example, GP population sizes run from ten to
a million members or more, but at present there is no practical guide
to knowing when to choose which size.\par

To continue addressing this issue, this paper builds on a previous
paper \cite{sastry:2003:GPTP} wherein we considered the building block
supply problem for GP.  In this earlier step, we asked what population
size is required to ensure the presence of all raw building blocks for
a given tree size (or size distribution) in the initial
population. The building-block supply based population size is
conservative because it does not guarantee the growth in the market
share of good substructures. That is, while ensuring the
building-block supply is important for a selecto-recombinative
algorithm's success, ensuring a growth in the market share of good
building blocks by correctly deciding between competing building
blocks is also critical \cite{Goldberg:2002:DOI}. Furthermore, the
population sizing for GA success is usually bounded by the population
size required for making good decisions between competing building
blocks. Our results herein show this to be the case, at least for the
{\tt ORDER} problem.\par

 Therefore, the purpose of this paper is to derive a population-sizing
model to ensure good decision making between competing building
blocks. Our analytical approach is similar to that used by
\citeN{Goldberg:1992:ComplexSystems} for developing a
population-sizing model based on decision-making for genetic
algorithms (GAs). In our population-sizing model, we incorporate
factors that are common to both GP and GAs, as well as those that are
unique to GP. We verify the populations-sizing model on three
different test problem that span the dimension of building block {\em
expression\/}---thus, modeling the phenomena of bloat at various
degrees. Using {\tt ORDER}, with {\tt UNITATION} as its fitness function,
provides a model problem where, per tree, a building block can be
expressed only once despite being present multiple times.  At the
opposite extreme, our {\tt LOUD} problem models a building block being
expressed each time it is present in the tree. In between, the {\tt
ON-OFF} problem provides tunability of building block expression. A
parameter controls the frequency with which a `function' can suppress
the expression of the subtrees below it, thus effecting how frequently
a tree expresses a building block. This series of experiments not only
validates the population-sizing relationship, but also empirically
illustrates the relationship between population size and problem
difficulty, solution complexity, bloat and tree structure.\par

We proceed as follows: The next section gives a brief overview of past
work in developing facetwise population-sizing models in both GAs and
GP. In Section~\ref{sec:ga-popsize}, we concisely review the
derivation by \cite{Goldberg:1992:ComplexSystems} of a population
sizing equation for GAs. Section~\ref{sec:defns} provides
GP-equivalent definitions of building blocks, competitions (a.k.a
partitions), trials, cardinality and building-block size. In
Section~\ref{sec:gp-popsize} we follow the logical steps of
\cite{Goldberg:1992:ComplexSystems} while factoring in GP perspectives
to derive a general GP population sizing equation. In
Section~\ref{sec:examples}, we derive and empirically verify the
population sizes for model problems that span the range variable BB
presence and its expressive probability. Finally,
section~\ref{sec:conclusions} summarizes the paper and provides key
conclusions of the study.

\section{Background}

%Added by Kumara
One of the key achievements of GA theory is the identification of the
building-block decision making to be a statistical one
\cite{Holland:1973:SIAM}.  \citeN{Holland:1973:SIAM} illustrated this
using a 2$^k$-armed bandit model.  Based on Holland's work,
\citeN{DeJong:1975:PhD} proposed equations for the 2-armed bandit
problem without using Holland's assumption of foresight. He recognized
the importance of noise in the decision-making process. He also
proposed a population-sizing model based on the signal and noise
characteristics of a problem.  \citeANP{DeJong:1975:PhD}'s suggestion
went unimplemented till the study by
\citeN{Goldberg:1991:ComplexSystems}.
\citeANP{Goldberg:1991:ComplexSystems} computed the fitness variance
using Walsh analysis and proposed a population-sizing model based on
the fitness variance.\par

A subsequent work \cite{Goldberg:1992:ComplexSystems} proposed an
estimate of the population size that controlled decision-making
errors. Their model was based on deciding correctly between the best
and the next best BB in a partition in the presence of noise arising
from adjoining BBs.  This noise is termed as {\em collateral noise}
\cite{Goldberg:1991:ComplexSystems}.  The model proposed by
\shortciteANP{Goldberg:1992:ComplexSystems} yielded practical
population-sizing bounds for selectorecombinative GAs. More recently
\citeN{Harik:1999:ECJ} refined the population-sizing model proposed by
\citeN{Goldberg:1992:ComplexSystems}. \shortciteANP{Harik:1999:ECJ}
proposed a tighter bound on the population size required for
selectorecombinative GAs.  They incorporated both the initial BB
supply model and the decision-making model in the population-sizing
relation. They also eliminated the requirement that only a successful
decision-making in the first generation results in the convergence to
the optimum. To eliminate this requirement, they modeled the
decision-making in subsequent generations using the well known
gambler's ruin model \cite{Feller:1970}. \citeN{Miller:1997:PhD} extended the
population-sizing model for noisy environments and
\citeN{CantuPaz:2000} applied it for parallel GAs.\par

While, population-sizing in genetic algorithms has been successfully
studied with the help of facetwise and dimensional models, similar
efforts in genetic programming are still in the early
stages. Recently, we developed a population sizing model to ensure the
presence of all raw building blocks in the initial population size. We
first derived the exact population size to ensure adequate supply for
a model problem named {\tt ORDER}. {\tt ORDER} has an expression
mechanism that models how a primitive in GP is expressed depending on
its spatial context.  We empirically validated our supply-driven
population size result for {\tt ORDER} under two different fitness
functions: {\tt UNITATION} where each primitive is a building block with
uniform fitness contribution, and {\tt DECEPTION} where each of $m$
subgroups, each subgroup consisting of $k$ primitives, has its fitness
computed using a deceptive trap function.\par

After dealing specifically with {\tt ORDER} in which, per tree, a building
block can be expressed at most once, we considered the general case of
ensuring an adequate building block supply where every building block
in a tree is always expressed. This is analogous to the instance of a
GP problem that exhibits no bloat. In this case, the supply equation
does not have to account for subtrees that are present yet do not
contribute to fitness. This supply-based population size equation is:
\begin{equation} \label{eqn:supply-ps}
n = \frac{1}{\sz{}}2^k\kappa\left(\log\kappa - \log\epsilon\right).
\end{equation}
where $\kappa$ enumerates the partition or building block competition,
$k$ is the building-block size, $\epsilon$ is supply error and \szm{}
is average tree size.\par

In the context of supply, to finally address the reality of bloat, we
noted that the combined probability of a building block being present
in the population and its probability of being expressed must be
computed and amalgamated into the supply derivation. This would imply
that Equation~\ref{eqn:supply-ps}, though conservative under the
assumed condition that every raw building block must be present in the
initial population, is an underestimate in terms of accounting for
bloat. Overall, the building block supply analysis yielded insight
into how two salient properties of GP: building block expression and
tree structure influence building block supply and thus influence
population size.  Building block expression manifests itself in `real
life' as the phenomena of bloat in GP. Average tree size in GP
typically increases as a result of the interaction of selection,
crossover and program degeneracy.

As a next step, this study derives a decision-making based
population-sizing model. We employ the methodology of
\citeN{Goldberg:1992:ComplexSystems} used for deriving a population
sizing relationship for GA. In this method, the population size is
chosen so that the population contains enough competing building
blocks that decisions between two building blocks can be made with a
pre-specified confidence. Compared to the GA derivation, there are two
significant differences. First, the collateral noise in fitness,
arises from a variable quantity of expressed BBs. Second, the number
of trials of a BB, rather than one per individual in the GA case,
depends on tree structure and whether a BB that is present in a tree
is expressed.  In the GP case, the variable, $\kappa$ related to
cardinality (e.g. the binary alphabet of a simple GA) and building
block defining length, is considerably larger because GP problems
typically use larger primitive sets. It is incorporated into the
relationship by considering BB expression and presence.\par

Before presenting the decision-making model for GP, we briefly discuss
the population-sizing model of \citeN{Goldberg:1992:ComplexSystems} in
the following section.

\section{GA Population Sizing from the Perspective of Competing
  Building Blocks}\label{sec:ga-popsize}

The derivational foundation for our GP population sizing equation is
the 1992 result for the selecto-recombinative GA by
\cite{Goldberg:1992:ComplexSystems} entitled ``Genetic Algorithms, Noise and the
Sizing of Populations''.  The paper considers how the GA can derive
accurate estimates of BB fitness in the presence of detrimental noise.
It recognizes that, while selection is the principal decision maker,
it distinguishes among individuals based on fitness and not by
considering BBs. Therefore, there is a possibility that an inferior BB
gets selected over a better BB in a competition due to noisy observed
contributions from adjoining BBs that are also engaged in competitions.

To derive a relation for the probability of deciding correctly between competing
BBs, the authors considered two individuals, one with the best BB and
the other with the second best BB in the same competition.
\cite{Goldberg:1992:ComplexSystems}.\par
\begin{figure}
\center
\epsfig{file=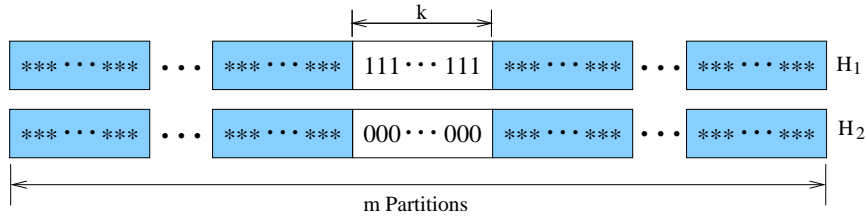,width=4.5in}
\caption[Illustration of competing building blocks.]{Two competing building blocks of size $k$, one is the best BB, $H_{1}$, and the other is the second best BB, $H_{2}$.}
\label{fig:bbdecision}
\end{figure}
\begin{figure}[h]
\center
\subfigure[Few samples]{\epsfig{file=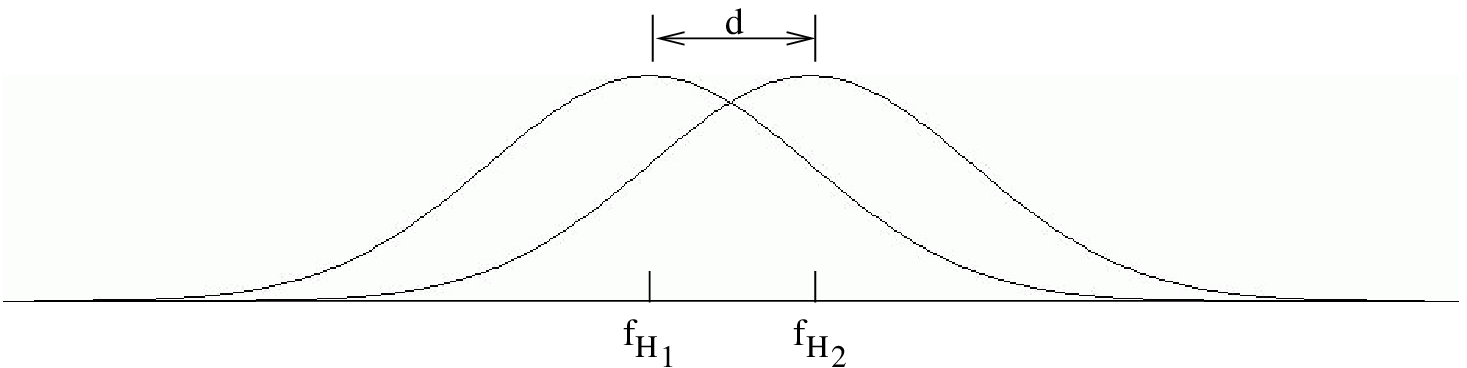,width=2.2in}}
\subfigure[Lots of samples]{\epsfig{file=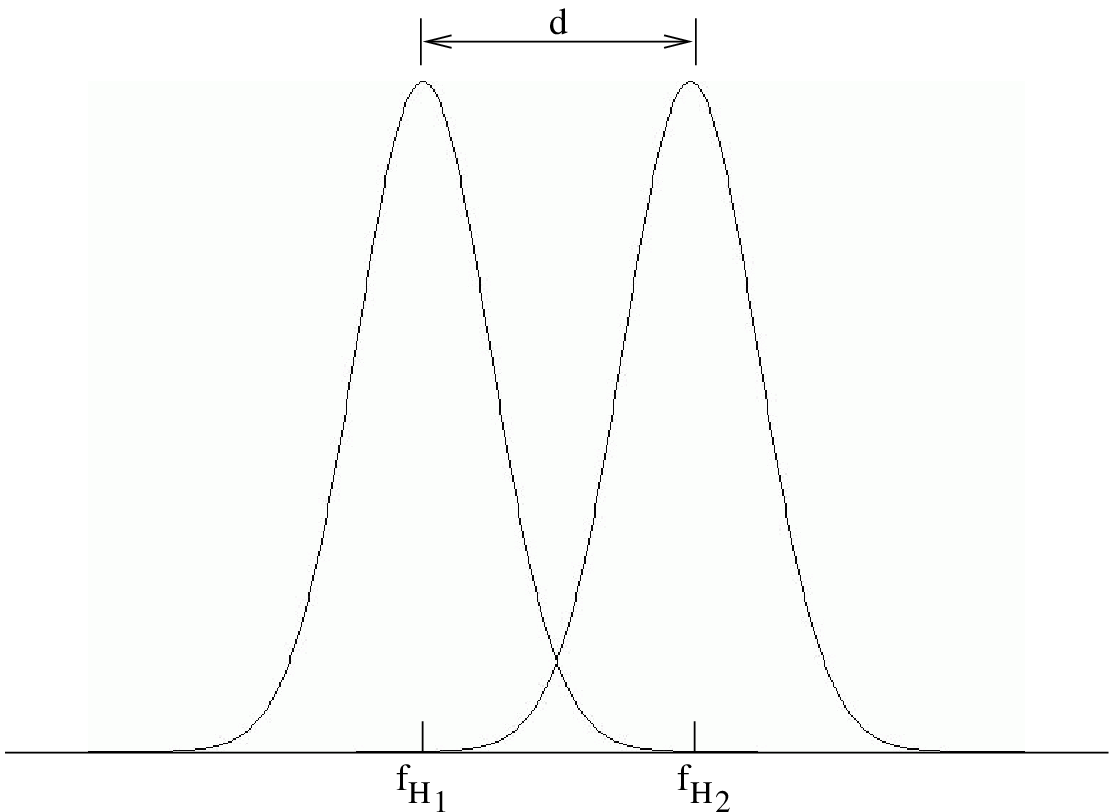,width=2.2in}}
%\subfigure[Few samples]{\includegraphics[width=2.2in]{dist1.pdf}}
%\subfigure[Lots of samples]{\includegraphics[width=2.2in]{dist2.pdf}}
\caption[Fitness distribution of competing building blocks.]{Fitness
  distribution of individuals in the population containing the two
  competing building blocks, the best BB $H_{1}$, and the second best
  BB $H_{2}$. When two mean fitness distributions overlap, low sampling
increases the likelihood of estimation error. When sampling around
each mean fitness is increased, fitness distributions are less likely to be
inaccurately estimated.}
\label{fig:bbfitdist}
\end{figure}

Let $i_1$ and $i_2$ be these two individuals with $m$ non-overlapping BBs of
size $k$ as shown in figure~\ref{fig:bbdecision}. Individual $i_1$ has
the best BB, $H_{1}$ ($111\cdots 111$ in figure~\ref{fig:bbdecision})
and individual $i_2$ has the second best BB, $H_{2}$ ($000\cdots 000$
in figure~\ref{fig:bbdecision}). The fitness values of $i_1$ and $i_2$
are $f_{H_1}$ and $f_{H_2}$ respectively. To derive the probability of
correct decision making, we have to first recognize that the fitness
distribution of the individuals containing $H_{1}$ and $H_{2}$ is
Gaussian since we have assumed an additive fitness
function and the central limit theorem applies. Two possible 
fitness distributions of individuals containing BBs $H_{1}$ and
$H_{2}$ are illustrated in figure~\ref{fig:bbfitdist}. 

The distance between the mean fitness of individuals containing $H_1$,
$\overline{f}_{H_1}$, and the mean fitness of individuals containing
$H_2$, $\overline{f}_{H_2}$, is the {\em signal}, $d$. That is
\begin{equation}
d = \overline{f}_{H_1} - \overline{f}_{H_2}.
\end{equation}

Recognize that the probability of correctly deciding between $H_1$ and
$H_2$ is equivalent to the probability that $f_{H_1}-f_{H_2} >
0$. Also, since $f_{H_1}$ and $f_{H_2}$ are normally distributed,
$f_{H_1}-f_{H_2}$ is also normally distributed with mean $d$ and
variance $\sigma^2_{H_1} + \sigma^2_{H_2}$, where $\sigma^2_{H_1}$ and
$\sigma^2_{H_2}$ are the fitness variances of individuals containing
$H_1$ and $H_2$ respectively. That is,
\begin{equation}
f_{H_1}-f_{H_2} \sim {\mathcal{N}}(d,\sigma^2_{H_1}+\sigma^2_{H_2}).
\end{equation}
The probability of correct decision making, $p_{dm}$, is then given by
the cumulative density function of a unit normal variate which is
the signal-to-noise ratio :
\begin{equation}
p_{dm} = \Phi\left({d \over \sqrt{\sigma^2_{H_1} + \sigma^2_{H_2}}}\right).
\end{equation}
Alternatively, the probability of making an error on a single
trial of each BB can estimated by finding the probability $\alpha$
such that 
\begin{equation}
z^2(\alpha) = \frac{d^2}{\sigma^2_{H_1} + \sigma^2_{H_2}}
\end{equation}
where $z(\alpha)$ is the ordinate of a unit, one-sided normal
deviate. Notationally $z(\alpha)$ is shortened to $z$.

Now, consider the BB variance, $\sigma^2_{H_1}$ (and
$\sigma^2_{H_2}$): since it is assumed the fitness function is the sum
of $m$ independent subfunctions each of size $k$, $\sigma^2_{H_1}$
(and similarly $\sigma^2_{H_2}$) is the sum of the variance of the
adjoining $m-1$ subfunctions. Also, since it is assumed that the $m$
partitions are uniformly scaled, the variance of each subfunction is
equal to the average BB variance, $\sigma^2_{bb}$. Therefore,
\begin{equation}
\mbox{GA BB Variance:\hspace{0.25in}} 
\sigma^2_{H_1} = \sigma^2_{H_2} = (m-1)\sigma^2_{bb}.\label{eqn:ga-bb-variance}
\end{equation}
A population-sizing equation was derived from this error probability
by recognizing that as the number of trials, $\trials{}$, increases, the variance of
the fitness is decreased by a factor equal to the trial quantity:
\begin{equation}
z^2(\alpha) = \frac{d^2}{\frac{(m-1)\sigma_{bb}}{\trials{}}}
\end{equation} 

To derive the quantity of trials, $\trials{}$, assume a uniformly
random population (of size $n$).  Let $\chi$ represent the cardinality
of the alphabet (2 for the GA) and $k$ the building-block size. For
any individual, the probability of $H_1$ is $1/\kappa$ where $\kappa =
\chi^{k}$. There is exactly one instance per individual of the
competition, $\phi = 1$. Thus,

\begin{equation}\label{eqn:trials}
\trials{} = n \cdot p_{BB} \cdot \phi = n \cdot 1/\kappa \cdot 1 = n/\kappa
\end{equation}

By rearrangement and calling $z^2$ the coefficient $c$
(still a function of $\alpha$) a fairly general population-sizing
relation was obtained:
\begin{equation}
n = 2c\chi^k (m-1)\frac{\sigma^2_{bb}}{d^2}\label{eqn:ga-popsize}
\end{equation}
To summarize, the decision-making based population sizing model in GAs
consists of the following factors:
\begin{itemize}
\item {\bf Competition complexity}, quantified by the total number of competing building blocks, $\chi^k$. 
\item {\bf Subcomponent Complexity}, quantified by the number of building blocks, $m$.
\item {\bf Ease of decision making}, quantified by the signal-to-noise ratio, $d/\sigma^2_{bb}$.
\item {\bf Probabilistic safety factor}, quantified by the coefficient $c$.
\end{itemize}

\section{GP Definitions for a Population Sizing
  Derivation}\label{sec:defns} 

Most GP implementations reported in the literature use parse trees to
represent candidate programs in the population \cite{langdon:fogp}. We
have assumed this representation in our analysis. To simplify the
analysis further, we consider the following:
\begin{enumerate}
\item A primitive set of the GP tree is $\mathcal{F} \cup \mathcal{T}$
where $\mathcal{F}$ denotes the set of functions (interior nodes to a
GP parse tree) and $\mathcal{T}$ denotes the set of terminals (leaf
nodes in a GP parse tree). 
\item The cardinality of $\mathcal{F} = \chi_{f}$ and the cardinality
of $\mathcal{T} =\chi_{t}$.
\item The arity of all functions in the primitive set is two: All
functions are binary and thus the GP parse trees generated from the
primitive set are binary.
\end{enumerate}
We believe that our analysis could be extended to primitive sets
containing functions with arity greater than two (non-binary trees).
We also note that our assumption closely matches a common GP
benchmark, symbolic regression, which frequently has arithmetic
functions of arity two.

\begin{figure}[hbt]
\centering
\mbox{\subfigure[]{\epsfig{file=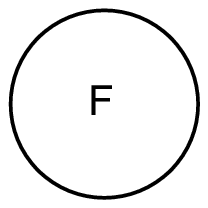,width=1cm}}%%\hspace{0.8cm}
      \subfigure[]{\epsfig{file=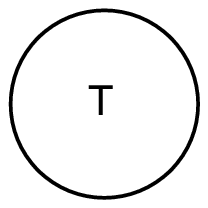,width=1cm}}%%\hspace{0.8cm}
      \subfigure[]{\epsfig{file=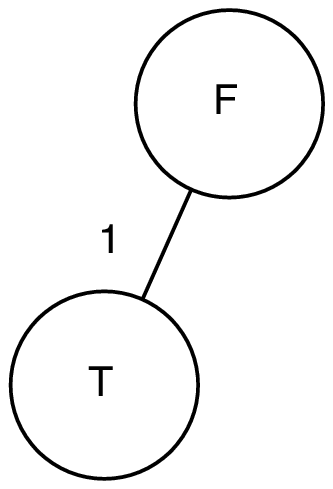,width=1.5cm}}
      \subfigure[]{\epsfig{file=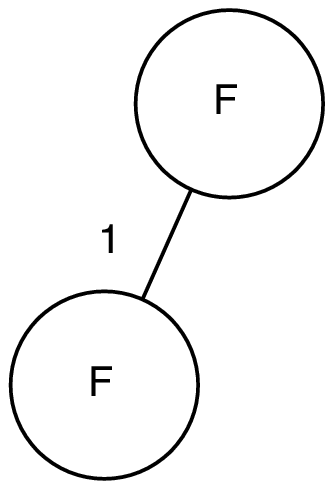,width=1.5cm}}
      \subfigure[]{\epsfig{file=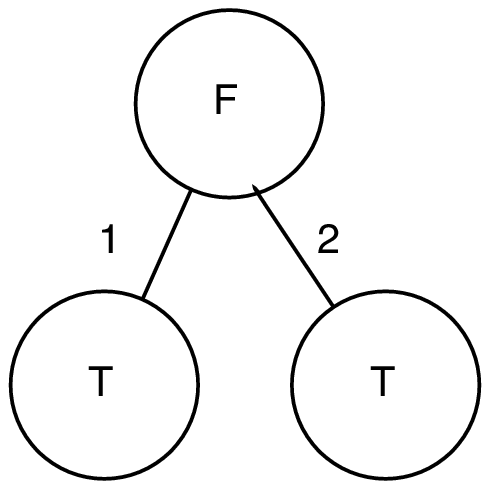,width=2cm}}
      \subfigure[]{\epsfig{file=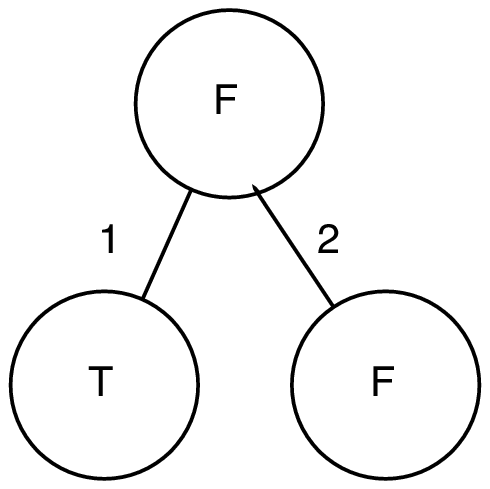,width=2cm}}
      \subfigure[]{\epsfig{file=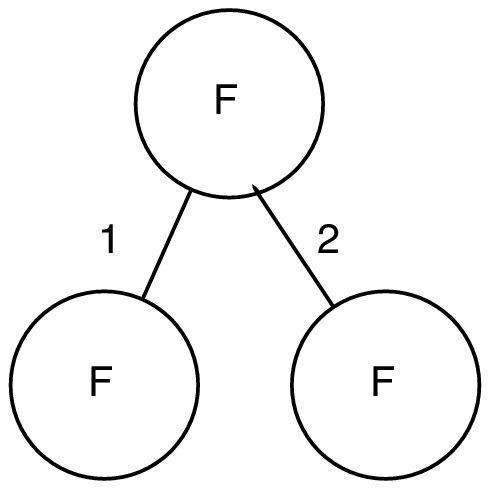,width=2cm}}}
%\mbox{\subfigure[]{\includegraphics[width=1cm]{f.pdf}} 
%           \subfigure[]{\includegraphics[width=1cm]{t.pdf}}      
%           \subfigure[]{\includegraphics[width=1.5cm]{ft1.pdf}}      
%           \subfigure[]{\includegraphics[width=1.5cm]{ff1.pdf}}      
%           \subfigure[]{\includegraphics[width=2cm]{ftt.pdf}}      
%           \subfigure[]{\includegraphics[width=2cm]{ftf.pdf}}      
%           \subfigure[]{\includegraphics[width=2cm]{fff.pdf}}      
%}     
\caption{The smallest tree fragments in GP. Fragments (c) and (d) have
mirrors where the child is 2nd parameter of the function.  Likewise,
fragment (f) has mirror where 1st and 2nd parameters of the function
are reversed.Recall that a tree fragment is a similarity template:
based on the similarity it defines, it also defines a competition. A
tree fragment, in other words, is a competition. (At other times we
have also used the term {\it partition} interchangeably with tree
fragment or competition)} \label{fig:partitions}
\end{figure}

As in our BB supply paper \cite{sastry:2003:GPTP}, our analysis adopts a
definition of a GP schema (or similarity template) called a ``tree
fragment''.  A tree fragment is a tree with at least one leaf that is
a ``don't care'' symbol. This ``don't care'' symbol can be matched by
any subtree (including degenerate leaf only trees).  As before, we are
most interested in only the small set set of tree fragments that are
defined by three or fewer nodes. See Figure~\ref{fig:partitions} for
this set.

The defining length of a tree fragment is the sum of its quantities of
function symbols, $\mathcal{F}$, and terminal symbols, $\mathcal{T}$:
\begin{equation}
k = N_f + N_t
\end{equation}
Because a tree fragment is a similarity template, it also represents
a competition. Since this paper is concerned
with decision making, we will therefore use ``competition'' instead
of a ``tree fragment''.  The size of a competition (i.e. how many
BBs compete) is
\begin{equation}
 \kappa = \chi_{f}^{N_f}*\chi_{t}^{N_t}
 \label{NumCompetSchema}
\end{equation}
As mentioned in \cite{sastry:2003:GPTP}, because a tree fragment is
defined without any positional anchoring, it can appear multiple times
in a single tree. We denote the number of instances of a tree fragment
that are present in a tree of size $\sz{}$, (a.k.a the quantity of a tree fragment in
a tree) as $\phi$. This is equivalent to the instances of a
competition as $\phi$ is used in the GA case (see Equation~\ref{eqn:trials}).
For full binary trees:
\begin{equation}
\phi \approx 2^{-k}\sz{}
\end{equation}
Later, we will explain how $\phi$ describes {\it potential} quantity of
 per tree'' of a BB.

\section{GP Population Sizing based on Decision Making} \label{sec:gp-popsize}

We now proceed to derive a GP population sizing relationship based on
building block decision making. Preliminarily, unless noted, we make
the same assumptions as the GA derivation of
Section~\ref{sec:ga-popsize}.

The first way the GP population size derivation diverges from the GA
case is how BB fitness variance (i.e. $\sigma^2_{H_1}$ and
$\sigma^2_{H_2}$) is estimated (for reference, see
Equation~\ref{eqn:ga-bb-variance}). Recall that for the GA the source of a BB's 
fitness variance was collateral noise from the $ (m-1)$ competitions
of its adjoining BBs. In GP, the source of collateral noise is the average number of
adjoining BBs present
and expressed in each tree, denoted as $\bar{q}$. Thus:
\begin{equation}
\mbox{GP BB Variance:\hspace{0.25in}}
\sigma^2_{H_1} = \sigma^2_{H_2} = [\bar{q}_{BB}^{expr}(m,\sz{}) -1]\sigma^2_{bb}.
\label{eqn:gp-bb-variance}
\end{equation}

Thus, the probability of making an error on a single trial of the BB
can be estimated by finding the probability $\alpha$ such that 
\begin{equation}
z^2(\alpha) = \frac{d^2}{2[\bar{q}_{BB}^{expr} -1]\sigma^2_{bb}}
\end{equation}

The second way the GP population size derivation diverges from the GA
case is in how the number of trials of a BB is estimated (for
reference, see Equation~\ref{eqn:trials}).  As with the GA, for GP we
assume a uniformly distributed population of size $n$.  In GP the
probability of a trial of a particular BB must account for it being
both present, $1/\kappa$, {\it and} expressed in an individual (or
tree), which we denote as $p_{BB}^{expr}$.  So, in GP:
\begin{equation}\label{eqn:gp-alpha}
\trials{} = {1\over\kappa} \cdot p_{BB}^{expr} \cdot \phi \cdot n
\end{equation}

Thus, the population size relationship for GP is:
\begin{equation}
n = 2c\frac{\sigma^2_{bb}}{d^2}\kappa\left[\bar{q}_{BB}^{expr} -1\right]
      \frac{1}{p_{BB}^{expr}\phi}
\label{eqn:gp-popsize}
\end{equation}
where $c = z^2(\alpha)$ is the square of the ordinate of a one-sided
standard Gaussian deviate at a specified error probability $\alpha$.
For low error values, $c$ can be obtained by the usual approximation
for the tail of a Gaussian distribution: $\alpha \approx
\exp(-c/2)/(\sqrt{2c})$.\par

Obviously, it is not always possible to factor the real-world problems
in the terms of this population sizing model.  A practical approach would first approximate
 $\phi = 2^{-k}(\sz{})$ trials per tree (the full binary tree assumption).  Then,
 estimate the size of the shortest program that will solve
the problem, (one might regard this as the Kolomogorov complexity of
the problem, $\sz{}_k$), and choose a multiple of this for $\sz{}$
in the model. In this case, $\bar{q} =  c_k m_k$. To ensure an initial supply of building blocks that is
sufficient to solve the problem, the initial population should be
initialized with trees of size $\sz{}$. Therefore, the
population sizing in this case can be written as
\begin{equation}
n = c\frac{\sigma^2_{bb}}{d^2}\kappa{\left(c_k m_k -1\right)2^{k+1} \over p_{BB}^{expr}\sz{}}
\label{eqn:gp-popsize1}
\end{equation}

Similar to the GA population sizing model, the decision-making based
population sizing model in GP consists of the following factors:
\begin{itemize}
\item {\bf Competition complexity}, quantified by
the total number of competing building blocks, $\kappa$.
\item {\bf Ease of decision making}, quantified
by the signal-to-noise ratio, $d/\sigma^2_{bb}$.
\item {\bf Probabilistic safety factor}, quantified by the coefficient $c$.
\item {\bf Number of subcomponents}, which unlike GA
population-sizing, depends not only on the minimum number of building
blocks, required to solve the problem $m_k$, but also tree size $\sz{}$, the size of the problem primitive set and how bloat factors into trees.
(quantified by $p_{BB}^{expr}$).
\end{itemize}

%Additionally one might execute a few experimental runs to assess the extent of bloat in solutions and use that to roughly estimate $p_{BB}^{expr}$.  
 
\section{Sizing Model Problems}\label{sec:examples}
This section derives the components of the population-sizing model
(Equation~\ref{eqn:gp-popsize}) for three test problems, {\tt ORDER},
{\tt LOUD}, and {\tt ON-OFF}. We develop the population-sizing
equation for each of theses problems and verify them with empirical
results. In all experiments we assume that $\alpha = 1/m$ and thus derive $c$. Table~\ref{tab:mandc} shows some of these values.  For all empirical experiments the 
the initial population is randomly generated with either full trees
or by the ramped half-and-half method.  The trees were allowed to grow up to a
maximum size of 1024 nodes. We used a tournament selection with
tournament size of 4 in obtaining the empirical results. We used
subtree crossover with a crossover probability of 1.0 and retained 5\%
of the best individuals from the previous population. A GP run was
terminated when either the best individual was obtained or when a
predetermined number of generations were exceeded. The average number
of BBs correctly converged in the best individuals were computed over
50 independent runs. The minimum population size required such that
$m-1$ BBs converge to the correct value is determined by a bisection
method \cite{Sastry:2002:Masters}. That is the error tolerance, $\alpha =
1/m$. The results of population size and convergence time was averaged
over 30 such bisection runs, while the results for the number of
function evaluations was averaged over 1500 independent runs.  We start with population sizing for {\tt ORDER}, where a
building block can be expressed at most once in a tree. 

%\begin{table}[htdp]
%\begin{center}
%\begin{tabular}{|c|c|}\hline
%$m$ & $c$ \\ \hline
%8 & .97 \\
%16 & 1.76 \\
%32 & 2.71 \\
%64 & 3.77 \\
%128 & 4.89 \\ \hline
%\end{tabular}
%\end{center}
%\label{tab:mandc}
%\caption{Values of $c = z^2(\alpha)$ used in population sizing equation.}
%\end{table}%

\begin{table}[htdp]
\center
\begin{tabular}{|c|c|c|c|c|c|}\hline
$m$ & 8  & 16   & 32   & 64   & 128\\\hline
$c$ &.97 & 1.76 & 2.71 & 3.77 & 4.89\\\hline
\end{tabular}
\label{tab:mandc}
\caption{Values of $c = z^2(\alpha)$ used in population sizing equation.}
\end{table}%

\subsection{{\tt ORDER}: At most one expression per building block per
      tree}
\begin{figure}[t]
\center
\epsfig{file=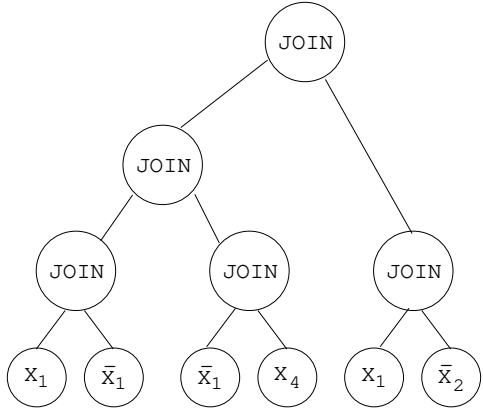, width=2.5in} 
\caption{A candidate solution for a 4-primitive {\tt ORDER}
  problem. The output of the program is $\{X_1,\bar{X}_2,X_3\}$ and
  its fitness is 2.}
\label{fig:treeOrder}
\end{figure}

{\tt ORDER} is a simple, yet intuitive expression mechanism which
makes it amenable to analysis and modeling
\cite{goldberg:1998:good,oreilly:1998:fssaGP}. The primitive set of
{\tt ORDER} consists of the primitive {\tt JOIN} of arity two and
complimentary primitive pairs $\left(X_i,\bar{X}_i\right)$, $i =
0,1,\cdots,m$ of arity one.  A candidate solution of the {\tt
ORDER} problem is a binary tree with {\tt JOIN} primitive at the
internal nodes and either $X_i$'s or $\bar{X}_i$'s at its leaves. The
candidate solution's expression is determined by parsing the program
tree inorder (from left to right).  The program expresses the value
$X_i$ if, during the inorder parse, a $X_i$ leaf is encountered before
its complement $\bar{X}_i$. Furthermore, only unique primitives are
expressed in {\tt ORDER} during the inorder parse. \par

For each $X_i$ (or $\bar{X}_i$) that is expressed, an equal unit of
fitness value is accredited. That is,
\begin{equation}
f_1(x_i) = \left\{ \begin{array}{ll} 1& {\mathrm{if}}~x_i \in
 \{X_1,X_2,\cdots,X_m\}\\
0 & otherwise\end{array} \right. .
\end{equation}
The fitness function for {\tt ORDER} is then defined as
\begin{equation}
F({\mathbf{x}}) = \sum_{i = 1}^{m} f_1\left(x_i\right),
\end{equation}
where ${\mathbf{x}}$ is the set of primitives expressed by the tree.
The output for optimal solution of a $2m$-primitive {\tt ORDER}
problem is $\{X_1,X_2,\cdots,X_m\}$, and its fitness value is
$m$. The building blocks in {\tt ORDER} are the primitives, $X_i$,
that are part of the subfunctions that reduce error (alternatively
improve fitness). The shortest perfect program is $\sz{}_k=2m-1$.\par

For example, consider a candidate solution for a 4-primitive {\tt
  ORDER} problem as shown in figure~\ref{fig:treeOrder}. The sequence
  of leaves for the tree is $\{X_1,$ $\bar{X}_1,$ $\bar{X}_1,$ $X_4,$
  $X_1,$ $\bar{X}_2\}$, the expression during inorder parse is
  $\{X_1,$ $\bar{X}_2,$ $X_4\}$, and its fitness is 2. For more
  details, motivations, and analysis of the {\tt ORDER} problem, the
  interested reader should refer elsewhere
  \cite{goldberg:1998:good,oreilly:1998:fssaGP}.\par

For the {\tt ORDER} problem, we can easily see that $\sigma^2_{bb} =
0.25$, $d = 1$, and $\phi = 1$. From \citeN{sastry:2003:GPTP}, we know that
\begin{equation}
\label{eqn:pbbExpOrder}p_{BB}^{expr} \approx \exp\left[-k\cdot e^{-{\sz{}\over 2m}}\right].
\end{equation}

Additionally, for {\tt ORDER}, $\bar{q}_{BB}^{expr}$ is given by
\begin{equation}
\label{eqn:nbbExpOrder} \bar{q}_{BB}^{expr} = 1 + \sum_{i =
  0}^{m-1}\left(\begin{array}{c}m-1\\i\end{array}\right) i \sum_{j =
  0}^{i}\left(\begin{array}{c}i\\j\end{array}\right)(-1)^j\left({i-j+1
  \over m}\right)^{n_l-1},
\end{equation}
where, $n_l$ is the average number of leaf nodes per tree in the
population. The derivation of the above equation was involved and detailed. It is provided in Appendix~\ref{sec:orderDERIVATION}). 

Substituting the above relations (Equations~\ref{eqn:pbbExpOrder}
and \ref{eqn:nbbExpOrder}) in the population-sizing model
(Equation~\ref{eqn:gp-popsize}) we obtain the following population-sizing
equation for {\tt ORDER}:
\begin{equation}
\label{eqn:popSizeOrder}
n = 2^{k-1} z^2(\alpha)\left({\sigma^2_{bb} \over
  d^2}\right)\left[\bar{q}_{BB}^{expr}
  -1\right]\exp\left[k\cdot e^{-{\sz{}\over 2m}}\right].
\end{equation}
\begin{figure}[tbh]
\center
\epsfig{file=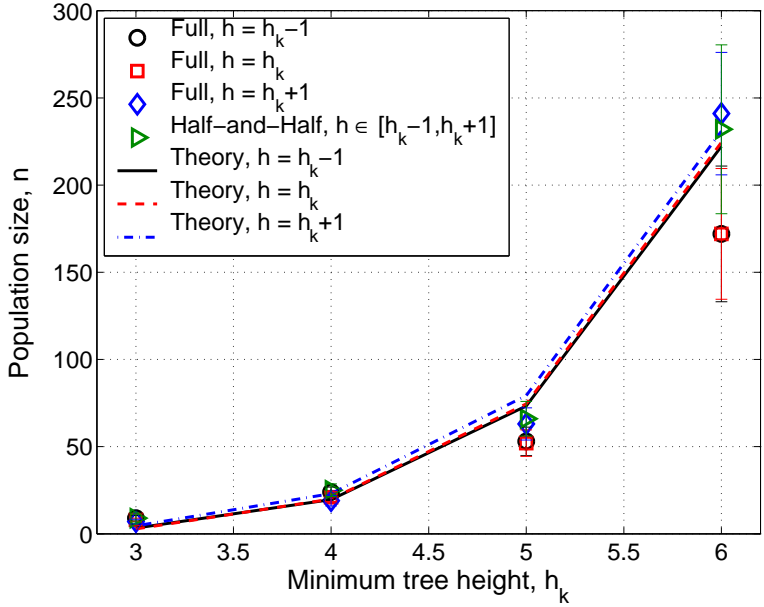, width=4in} 
\caption{Empirical validation of the population-sizing model
  (Equation~\ref{eqn:popSizeOrder}) for {\tt{ORDER}} problem.  Tree height $h_k$  equals $2^m$ and $\sz{} = 2m -1 = 2^{h+1} -1$.}
\label{fig:OrderPopSize}
\end{figure}

\begin{figure}[tbh]
\center
\subfigure[Convergence Time, $t_c$]{\epsfig{file=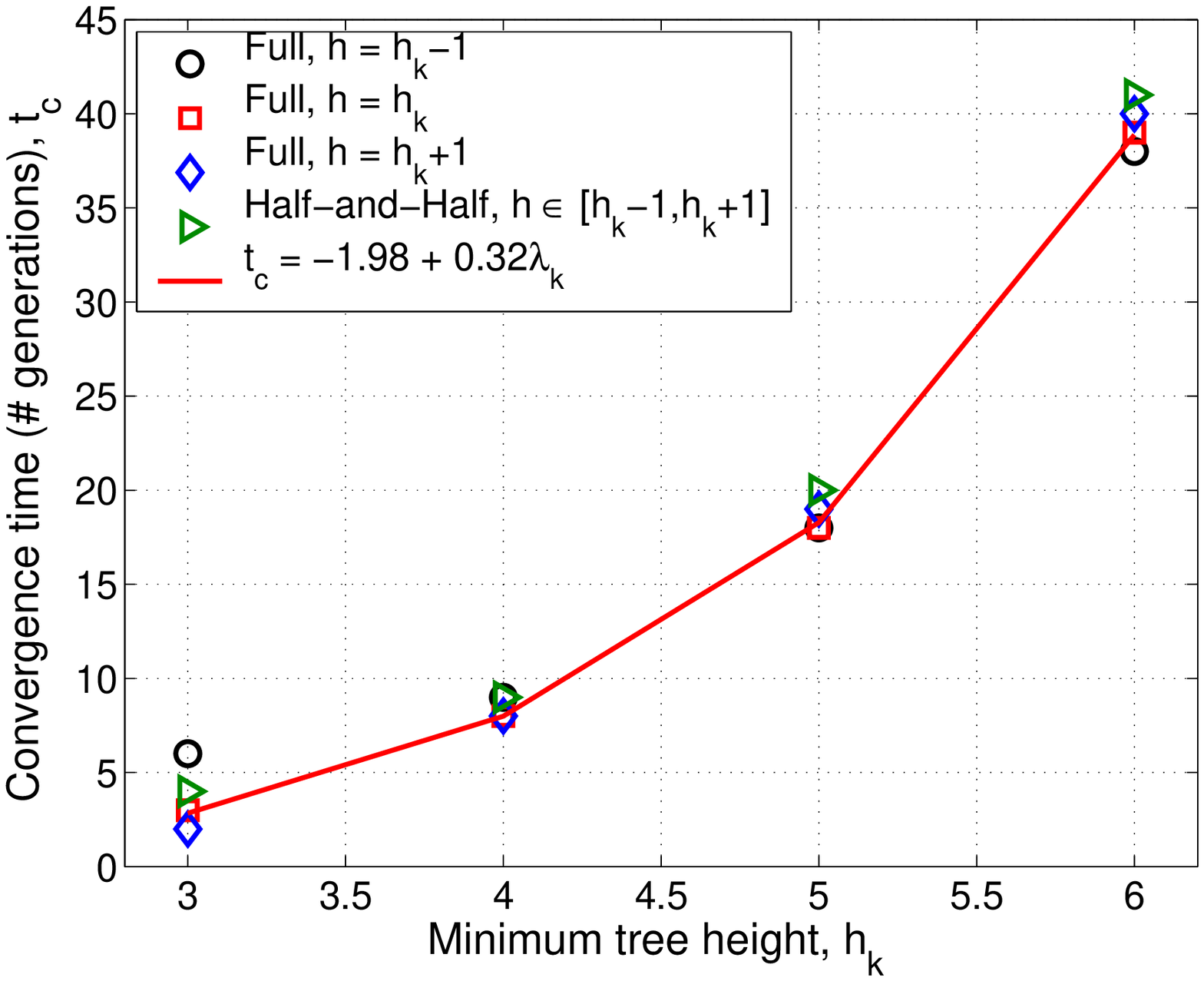, width=3in}}
\subfigure[Total number of function evaluations, $n_{fe}$]{\epsfig{file=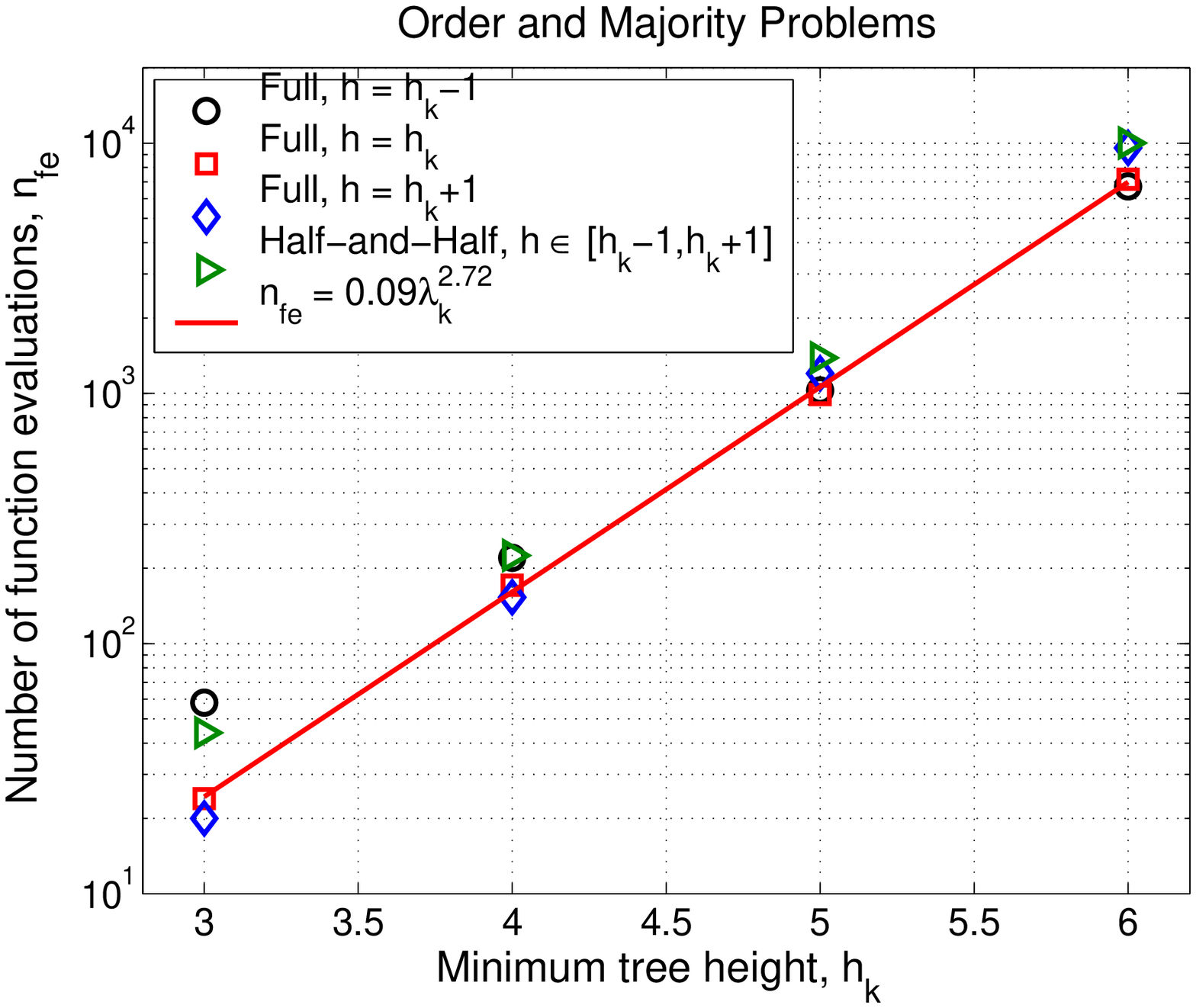, width=3in}}
%\subfigure[Convergence Time, $t_c$]{\includegraphics[width=3in]{orderConvTime.pdf}}
%\subfigure[Total number of function evaluations, $n_{fe}$]{\includegraphics[width=3in]{orderFuncEvals.pdf}}
\caption{Empirical results for the convergence time and the total
  number of function evaluations required to obtain the global
  solution for {\tt{ORDER}} problem. Note that $\sz_k = 2m -1$ so convergence time and the number of function evaluations scale linearly and cubically with the program size of the most compact solution or problem difficulty.  The implication is that population size for {\tt{ORDER}} problem is quadratic. }
\label{fig:OrderConvTime}
\end{figure}

The above population-sizing equation is verified with empirical
results in Figure~\ref{fig:OrderPopSize}. The 
initial population was randomly generated with either full trees
or by the ramped half-and-half method  with trees of heights, $h \in
[h_k-1,h_k+1]$, where, $h_k$ is the minimum tree height with an
average of $2m$ leaf nodes. \par

As shown in Figure~\ref{fig:OrderConvTime}, we empirically observed
that the convergence time and the number of function evaluations scale
linearly and cubically with the program size of the most compact
solution, $\sz{}_k$, respectively. From this empirical observation,
we can deduce that the population size for {\tt ORDER} scales
quadratically with the program size of the most-compact solution. For
{\tt ORDER}, $\sz{}_k = 2m -1$.\par

To summarize for the {\tt ORDER} problem, where a building block is
expressed at most once per individual, the population size scales as $n =
{\mathcal{O}}\left(2^k\sz{}_k^2\right)$, the convergence time scales
as $t_c = {\mathcal{O}}\left(\sz{}_k\right)$, and the total number
of function evaluations required to obtain the optimal solution scales
as $n_{fe} = {\mathcal{O}}\left(2^k\sz{}_k^3\right)$.

\subsection{{\tt LOUD}: Every building block in a tree is expressed}
In {\tt ORDER}, a building block could be expressed at most once in a
tree, however, in many GP problems a building block can be expressed
multiple times in an individual. Indeed, an extreme case is when every
building block occurrence is expressed. One such problem is a modified
version of a test problem proposed by \citeN{soule:2002:EuroGP} (see also
\cite{soule:2002:GPEM,soule:2003:GPTP}), which we call as {\tt LOUD}.\par

In {\tt LOUD}, the primitive set consists of an ``add'' function of
arity two, and three constant terminal 0, 1 and 4. The objective is to
find an optimal number of fours and ones. That is, for an individual
with $i$ 4s and $j$ 1s, the fitness function is given by
\begin{equation}
F({\mathbf{x}}) = \left|i - m_4\right| + \left|j - m_1\right|
\end{equation}
Therefore, even though a zero is expressed it does not contribute to
fitness. Furthermore, a 4 or 1 is expressed each time it appears in
an individual and each occurrence contributes to the fitness value of
the individual. Moreover, the problem size, $m = m_4 + m_1$ and$\sz_k = 2m-1$ .\par

For the {\tt LOUD} problem the building blocks are ``4'' and ``1''. It
is easy to see that for {\tt LOUD}, $\sigma^2_{BB} = 0.25$, $d = 1$,
$\phi = \sz/2$, and $p_{BB}^{expr} = 1/3$. Furthermore, the average
number of building blocks expressed is given by $\bar{q}_{BB}^{expr} =
2n_l/3 \approx \sz{}/3$. Substituting these values in the
population-sizing model (Equation~\ref{eqn:gp-popsize}) we obtain
\begin{equation}
\label{eqn:loudPopSize}n = 2\cdot3^{k} z^2(\alpha)\left({\sigma^2_{bb} \over
  d^2}\right)\left[\frac{1}{3}\sz{} -1\right] \cdot \left({2\over \sz{}}\right).
\end{equation}

\begin{figure}[tbh]
\center
\subfigure[Population size, $n$]{\epsfig{file=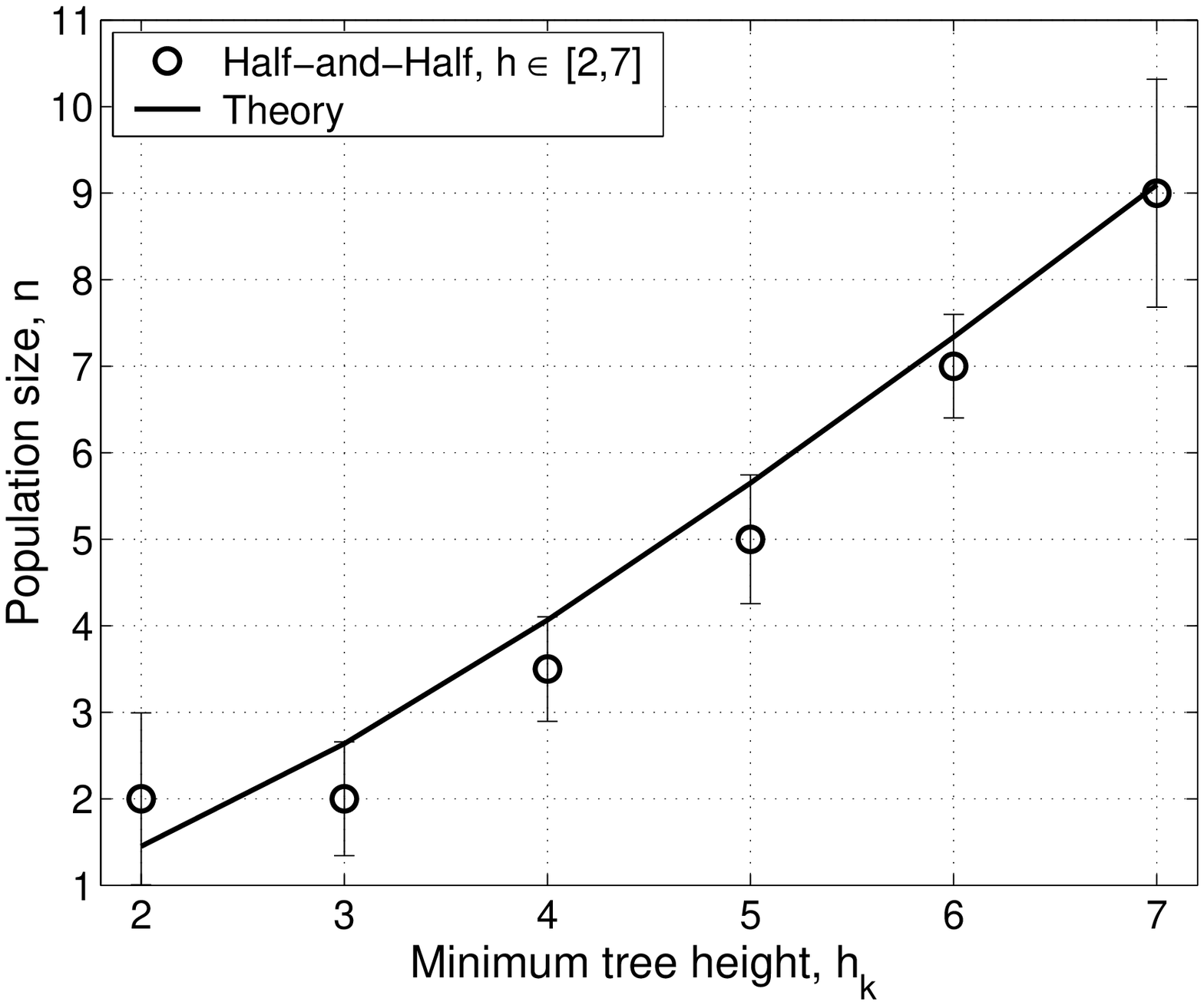, width=3in}}
\subfigure[Total number of function evaluations, $n_{fe}$]{\epsfig{file=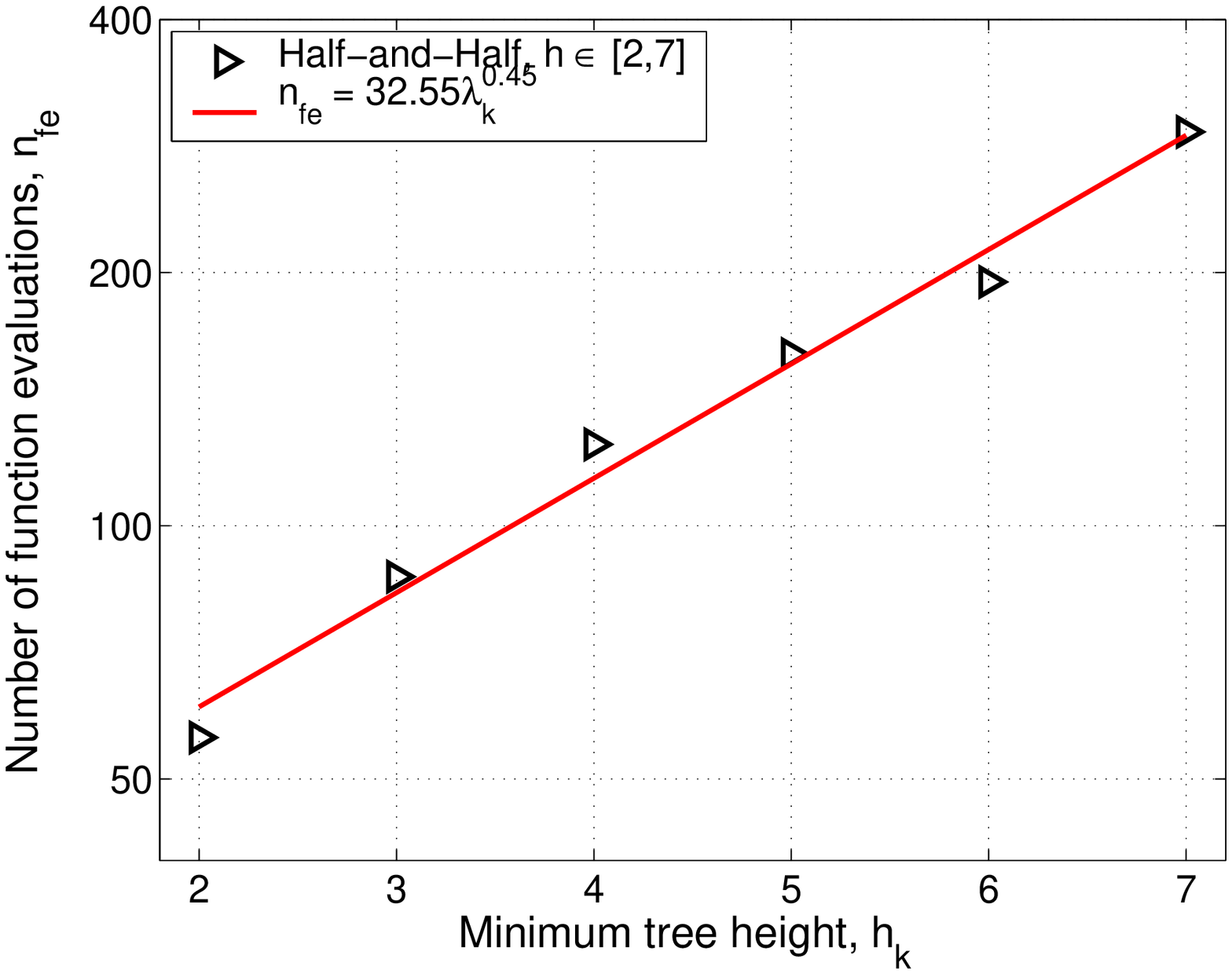, width=3in}}
%\subfigure[Population size, $n$]{\includegraphics[width=3in]{loudPopSize.pdf}}
%\subfigure[Total number of function evaluations, $n_{fe}$]{\includegraphics[width=3in]{loudFuncEvals.pdf}}
\caption{Empirical validation of the population-sizing model
  (Equation~\ref{eqn:loudPopSize}) and empirical results for the total
  number of function evaluations required to obtain the global
  solution for {\tt{LOUD}} problem.  Convergence time was constant with respect to problem size.  Note that $\sz_k = 2m -1$ so  the number of function evaluations scales sub-linearly with the program size of the most compact solution or problem difficulty.  The implication is that population size for {\tt{LOUD}} problem is sub-linear.}
\label{fig:loudPopSize}
\end{figure}

The above population-sizing equation is verified with empirical
results in Figure~\ref{fig:loudPopSize}. The initial population was randomly generated by the ramped
half-and-half method with trees of heights, $h \in [2,7]$ yielding an average tree size of 4.1 (this value is analytically 4.5). 

We empirically observed that the convergence time was constant with
respect to the problem size, and the number of function evaluations
scales sub-linearly with the program size of the most-compact
solution, $\sz{}_k$. From this empirical observation,
we can deduce that the population size for {\tt LOUD} scales
sub-linearly with the program size of the most-compact solution. For
{\tt LOUD} $\sz{}_k = 2m-1$.\par

To summarize for the {\tt LOUD} problem, where a building block is
expressed each time it occurs in an individual, the population size
scales as $n = {\mathcal{O}}\left(3^k\sz{}_k^{0.5}\right)$, the
convergence time is almost constant with the problem size, and, and
the total number of function evaluations required to obtain the
optimal solution scales as $n_{fe} =
{\mathcal{O}}\left(3^k\sz{}_k^{0.5}\right)$.

\subsection{{\tt ON-OFF}: Tunable building block expression}
In the previous sections we considered two extreme cases, one where a
building block could be expressed at most once in an individual and
the other where every building block occurrence is expressed. However,
usually in GP problems, some of the building blocks are expressed and
others are not. For example, a building block in a non-coded segment
is neither expressed nor contributes to the fitness. Empirically, \cite{luke:2000:cgnci} calculates the percentage of inviable nodes in runs  of the 6 and 11 bit multiplexer problems and symbolic regression over the course of a run. This value is seen to vary between problems and change over generations.  Therefore, the
third test function, which we call {\tt ON-OFF}, is one in which the
probability of a building block being expressed is tunable.\par

In {\tt ON-OFF}, the primitive set consists of two functions {\tt EXP}
and $\overline{{\mathtt{EXP}}}$ of arity two and terminal {\tt X}$_1$,
and {\tt X}$_2$. The function {\tt EXP} expresses its child nodes,
while $\overline{{\mathtt{EXP}}}$ suppresses its child nodes. Therefore
a leaf node is expressed only when all its parental nodes have the
primitive {\tt EXP}. This function can potentially approximate some
bloat scenarios of standard GP problems such as symbolic-regression
and multiplexer problems where invalidators are responsible for
nullifying a building block's effect \cite{luke:2000:cgnci}. The
probability of expressing a building block can be tuned by controlling
the frequency of selecting {\tt EXP} for an internal node in the
initial tree.\par

Similar to {\tt LOUD}, the objective for {\tt ON-OFF} is to find an
optimal number of fours and ones. That is, for an individual with $i$
{\tt X}$_1$s and $j$ {\tt X}$_2$s, the fitness function is given by
\begin{equation}
F({\mathbf{x}}) = \left|i - m_{X_1}\right| + \left|j - m_{X_2}\right|
\end{equation}
The problem size, $m = m_{X_1} + m_{X_2}$ and $\sz{}_k = 2m -1$.\par

\begin{figure}[t]
\center \epsfig{file=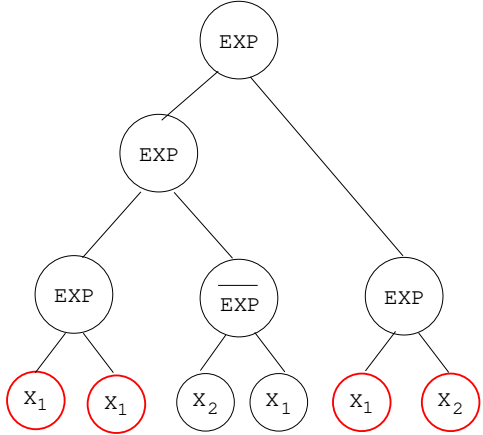, width=2.5in}
%\center\includegraphics[width=2.5in]{treeOnOff.pdf}
\caption{A candidate solution for a 2-primitive {\tt ON-OFF}
  problem. The output of the program is $\{X_1,X_1,X_1,X_2\}$ and
  its fitness is $|3-m_{x_1}| + |1 - m_{x_2}|$.}
\label{fig:treeOnOff}
\end{figure}

For example, consider a candidate solution for the {\tt LOUD} problem
  as shown in figure~\ref{fig:treeOnOff}. The terminals that are
  expressed are $\{X_1$, $X_1$, $X_1$, $X_2\}$ and the fitness is
  given by $|3-m_{x_1}| + |1 - m_{x_2}|$.\par

For the {\tt ON-OFF} problem the building blocks are $X_1$ and $X_2$,
$\sigma^2_{BB} = 0.25$, $d = 1$, $\phi = \sz{}/2$, and $p_{BB}^{expr} =
p_{EXP}^h$. Here, $p_{EXP}$ is the probability of a node being the
primitive {\tt EXP}. The average number of building blocks expressed
is given by $\bar{q}_{BB}^{expr} = n_l\cdot p_{EXP}^h \approx
\frac{s}{2}\cdot p_{EXP}^h$. Substituting these values in the population-sizing model
(Equation~\ref{eqn:gp-popsize}) we obtain
\begin{equation}
\label{eqn:onOffPopSize}n = 2^{k+1} z^2(\alpha)\left({\sigma^2_{bb} \over
  d^2}\right)\left[\frac{\sz{}}{2}p_{EXP}^h -1\right] \cdot \left({2\over \sz{} p_{EXP}^h}\right).
\end{equation}

\begin{figure}[tbh]
\center
\subfigure[Population size, $n$]{\epsfig{file=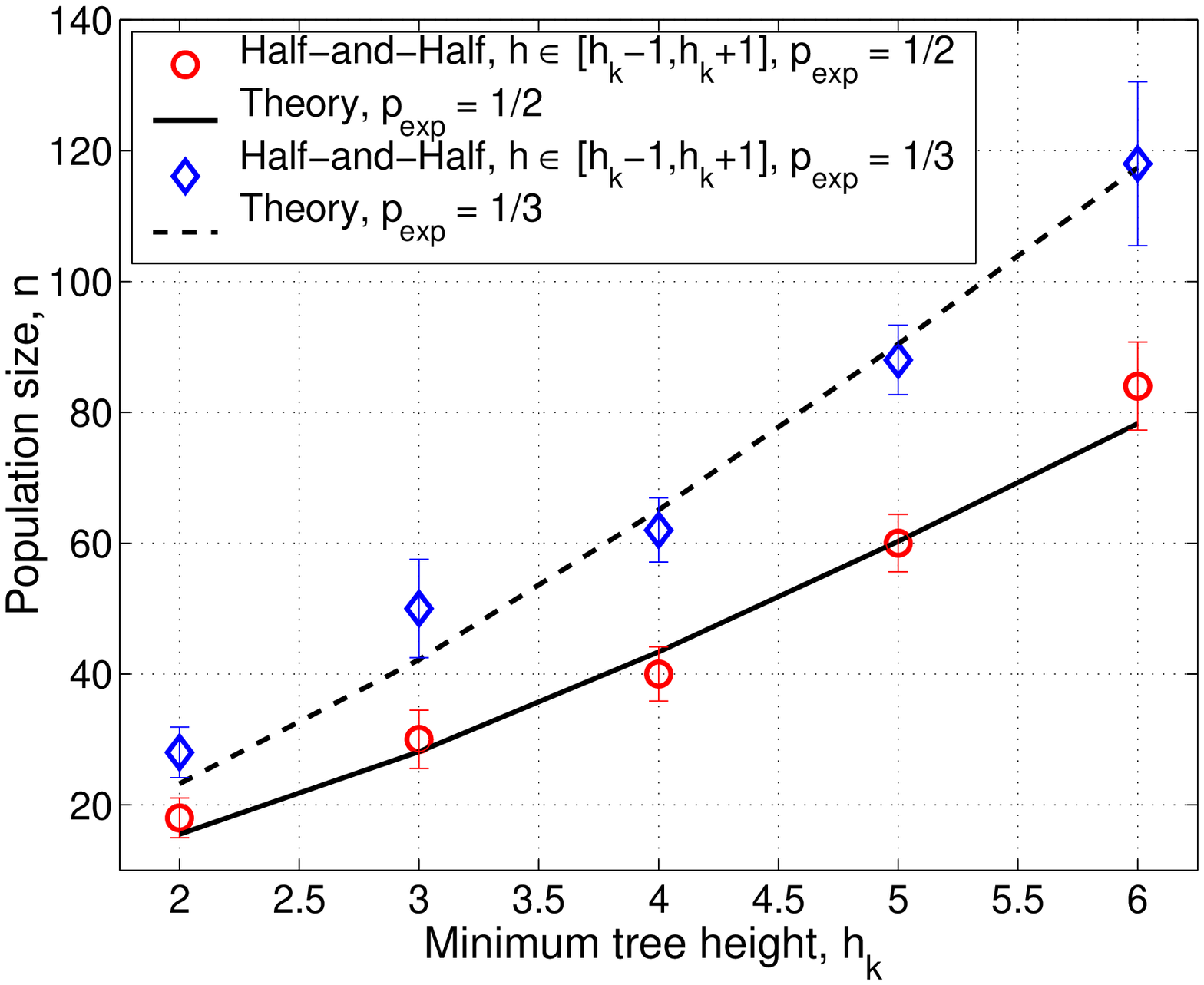, width=3in}}
\subfigure[Total number of function evaluations, $n_{fe}$]{\epsfig{file=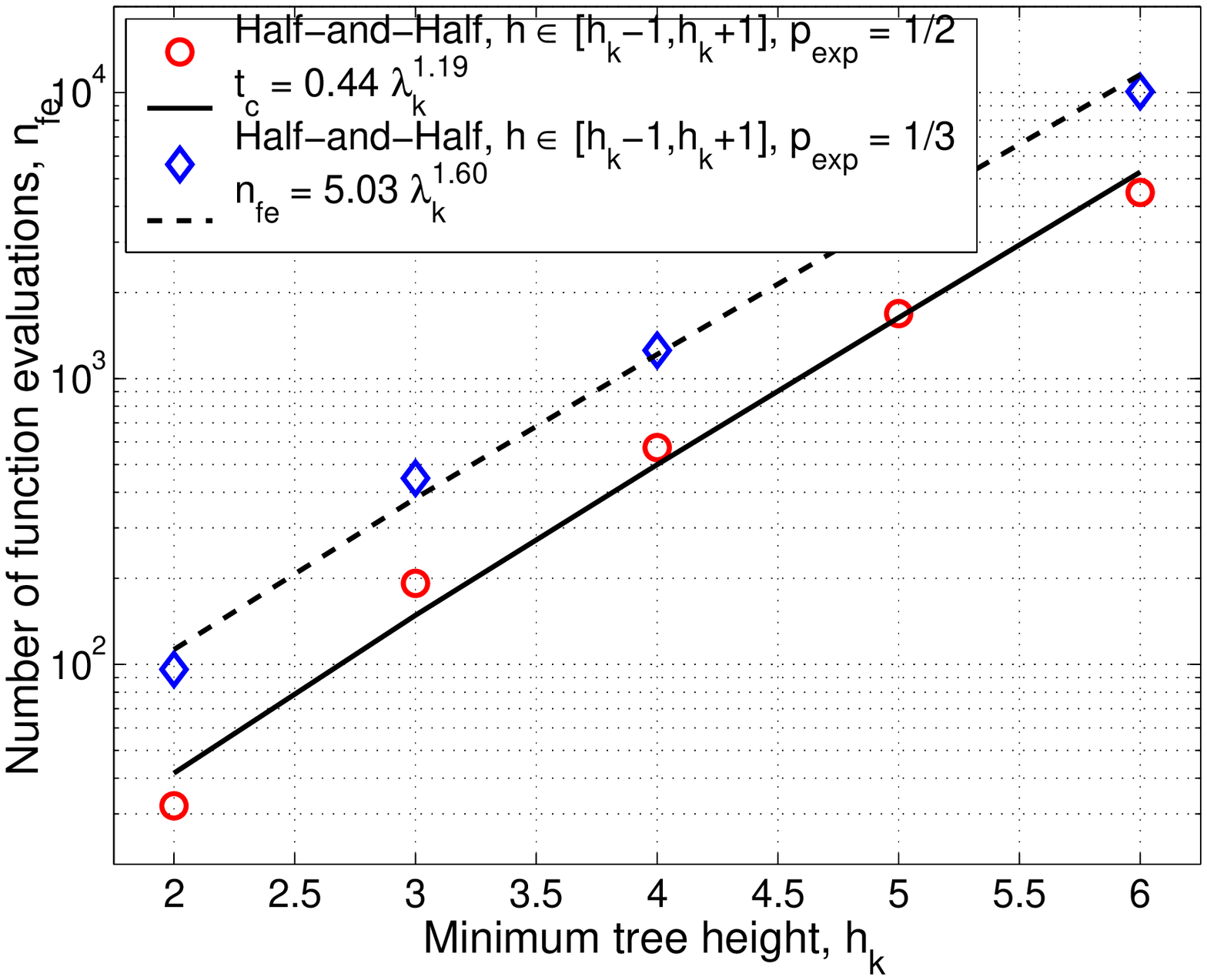, width=3in}}
%\subfigure[Population size, $n$]{\includegraphics[width=3in]{onOffPopSize.pdf}}
%\subfigure[Total number of function evaluations, $n_{fe}$]{\includegraphics[width=3in]{onOffFuncEvals.pdf}}
\caption{Empirical validation of the population-sizing model
  (Equation~\ref{eqn:onOffPopSize}) and empirical results for the total
  number of function evaluations required to obtain the global
  solution for {\tt{On-Off}} problem.  Convergence time was constant
  with respect to problem size. Note that $\sz_k = 2m -1$. The
  convergence time scales linearly ${\mathcal{O}}\left(\sz_k\right)$,
  and the number of function evaluations scales sub-quadratically
  ${\mathcal{O}}\left(\sz_k^1.5\right)$ with the program size of the
  most compact solution or problem difficulty. Therefore, the
  population size for {\tt{On-Off}} problem scales sub-linearly
  ${\mathcal{O}}\left(\sz_k^0.5\right)$.}
\label{fig:onOffPopSize}
\end{figure}

The above population-sizing equation is verified with empirical
results in Figure~\ref{fig:onOffPopSize}. The initial population was
randomly generated by the ramped half-and-half method with trees of
heights, $h \in [h_k-1,h_k+1]$, where $h_k$ is the minimum tree height
with an average of $m$ leaf nodes. We empirically observed that the
convergence time was linear with respect to the problem size, and
the number of function evaluations scales sub-quadratically with the
program size of the most-compact solution, $\sz{}_k$. From this
empirical observation, we can deduce that the population size for {\tt
On-Off} scales sub-linearly with the program size of the most-compact
solution ($\sz{}_k = 2m-1$).\par

To summarize for the {\tt On-Off} problem, where a building block
expression is tunable, the population size scales as $n =
{\mathcal{O}}\left(2^k\sz{}_k^{0.5}/p_{exp}\right)$, the convergence
time scales linearly as $t_c =
{\mathcal{O}}\left(2^k\sz{}_k/p_{exp}\right)$, and the total number of
function evaluations required to obtain the optimal solution scales as
$n_{fe} = {\mathcal{O}}\left(2^k\sz{}_k^{1.5}/p_{exp}^2\right)$.

\section{Conclusions}\label{sec:conclusions}

This contribution is a second step towards a reliable and accurate
model for sizing genetic programming populations. In the first step
the model estimated the minimum population size required to ensure
that every building block was present with a given certainty in the
initial population.  We accepted this conservative model (i.e. it
oversized the population) because in the process of deriving it, we
gained valuable insight into a) what makes GP different from a GA in
the sizing context and b) the implications of these differences. The
difference of GP's larger alphabet, while influential in implying GP
needs larger population sizes, was not a difficult factor to handle
while bloat and the variable length individuals in GP are more
complicated.

Moving to the second step, by considering a decision making model
(which is less conservative than the BB supply model), we extended the
GA decision making model along these dimensions: first, our model
retains a term describing collateral noise from competing BBs
($\bar{q}[m,\sz{}]$) but it recognizes that the quantity of these
competitors depends on tree size and the likelihood that the BB is
present and expresses itself (rather than behave as an
intron). Second, our model, like its GA counterpart, assumes that
trials decrease BB fitness variance, however, what was simple in a GA
-- there is one trial per population member, for the GP case is more
involved.  That is, the probability that a BB is present in a
population member depends both on the likelihood that it is present in
lieu of another BB {\it and} expresses itself, {\it plus} the number
of potential trials any BB has in each population member.

The model shows that, to ensure correct decision making within an
error tolerance, population size must go up as the probability of
error decreases, noise increases, alphabet cardinality increases, the
signal-to-noise ratio decreases {\it and} tree size decreases and
bloat frequency increases.  This obviously matches intuition.  There
is an interesting critical trade-off with tree size with respect to
determining population size: pressure for larger trees comes from the
need to express all correct BBs in the solution while pressure for
smaller trees comes from the need to reduce collateral noise from
competing BBs.

The model is conservative because ``it assumes that decisions are made
irrevocably during any given generation. It sizes the population to
ensure that the correct decision is made on average in a single
generation'' \cite{Goldberg:2002:DOI}.  In this way, it is similar to
the Schema Theorem.  A more accurate and different model would account
for how correct decision making accumulates over the course of a run,
and how, over the course of a run, improper decision making can be
rectified.

The fact that the model is based on statistical decision making means
that crossover does not have to be incorporated.  In GAs crossover
solely acts as a mixer or combiner of BBs. Interestingly, in GP,
crossover also interacts with selection with the potential result that
programs' size grows and structure changes.  When this happens, the
frequency of bloat can also change (see
\cite{luke:2000:cgnci,luke:dissertation} for examples of this with
multiplexer and symbolic regression).  These changes in size,
structure and bloat frequency imply a much more complex model if one
were to attempt to account for decision making throughout a run. They
also suggest that when using the model as a rule of thumb to size an
initial population it may prove more accurate if the practitioner
overestimates bloat in anticipation of subsequent tree growth causing
more than the bloat seen in the initial population, given its average
tree size.

It appears difficult to use this model with real problems where, among
the GP particular factors, the most compact solution and BB size is
not known and the extent of bloat can not be estimated. In the case of
the GA model, the estimation of model factors has been addressed by
\cite{Reed:2000:WRR}. They estimated variance with the
standard deviation of the fitness of a large random population. In the
GP case, this sampling population should be controlled for average
tree size. If a practitioner were willing to work with crude estimates
of bloat, BB size and most compact solution size, a multiple of the
size of the most compact solution could be plugged in, and bloat could
be used with that size to estimate the probability that a BB is
expressed and present and the average number of BBs of the same size
present and expressed, on average, in each tree.  In the future we
intend to experiment with the model and well known toy GP problems
(e.g. multiplexer, symbolic regression) where bloat frequency and most
compact problem size are obtainable, and simple choices for BB size
exist to see whether the ideal population size scales with problem
size within the order of complexity the model predicts.

Population sizing has been important to GAs and is
now important to GP, because it is the principle factor in controlling
ultimate solution quality.  Once the quality-size relation is
understood, populations can be sized to obtain a desired quality and
only two things can happen in empirical trials.  The quality goal can
be equaled or exceeded in which case, all is well with the design of
the algorithm, or (as is more likely) the quality target can be
missed, in which case there is some other obstacle to be overcome in
the algorithm design.  Moreover, once population size is understood in
this way it can be combined with an understanding of run duration (citation),
thereby yielding first estimates of GP run complexity, a key milestone
in making our understanding of these processes more rigorous.

\section*{Acknowledgments}
We gratefully acknowledge the organizers and reviewers of the 2004 GP
Theory and Practice Workshop.

This work was sponsored by the Air Force Office of Scientific
Research, Air Force Materiel Command, USAF, under grant F49620-00-0163
and F49620-03-1-0129, the National Science Foundation under ITR grant
DMR-99-76550 (at Materials Computation Center), and ITR grant
DMR-0121695 (at CPSD), and the Dept. of Energy under grant
DEFG02-91ER45439 (at Fredrick Seitz MRL). The U.S.  Government is
authorized to reproduce and distribute reprints for government
purposes notwithstanding any copyright notation thereon.\par

The views and conclusions contained herein are those of
the authors and should not be interpreted as necessarily representing
the official policies or endorsements, either expressed or implied, of
the Air Force Office of Scientific Research, the National Science
Foundation, or the U.S. Government.
\bibliographystyle{my-apa-uiuc}
\bibliography{GPbib,myBib}
\newpage
\appendix

\section{Derivation of the Average Number of Expressed Building Blocks
  for the {\tt{ORDER}} Problem}\label{sec:orderDERIVATION}
The following derivation provides expression for the average number of
expressed building blocks (BBs) (best or second best) in other partitions,
given that a best BB or second best BB is already expressed in a
particular partition. For example, I assume that either $X_1$ or
$\bar{X}_1$ is expressed in a tree. Therefore the total number of leaf
nodes available to potential express other BBs is $n_l - 1$.\par

Given that the problem has $m$ building blocks, the total number of
terminals, $\chi_t = 2m$ (Recall that the terminal set, ${\mathcal{T}}
\equiv \{X_1, \bar{X}_1, X_2, \bar{X}_2, \cdots, X_m,
\bar{X}_m\}$). Therefore, the total possible terminal sequences, given
$n_l - 1$ leaf nodes, $N_{{\mathrm{tot}}}$, is
\begin{equation}
N_{{\mathrm{tot}}} = \left(2m\right)^{n_l-1}.
\end{equation}

The number of building blocks that expressed in $n_l -1$ nodes vary
from 0 to $m-1$ (note that we assume that one building block is
already expressed). That is, if either $X_1$ or $\bar{X}_1$ are
present in the remaining $n_l-1$ leaf nodes, the number of expressed
building blocks other than $X_1$ or $\bar{X}_1$ is zero. Similarly if
there is at least one copy of one of the $m-1$ complementary
primitives present in $n_l-1$ leaf nodes, then the number of BBs
expressed other than $X_1$ or $\bar{X}_1$ is $m-1$. For brevity, in
the reminder of this report, the number of expressed BBs refer to only
the BBs expressed in $n_l-1$ leaf nodes.

Before proceeding with the derivation itself, we develop few
identities that will be used throughout the derivation.
\begin{equation}
\label{eqn:identity1} \sum_{j = 0}^{n}
\left(\begin{array}{c}n\\j\end{array}\right) = 2^n
\end{equation}

\begin{eqnarray}
\nonumber
\sum_{j=0}^{n}\left(\begin{array}{c}n\\j\end{array}\right)a^{n-j} &=&
a^n\sum_{j=0}^{n}\left(\begin{array}{c}n\\j\end{array}\right)\left({1\over
    a}\right)^j\\
\nonumber &=&
a^n\sum_{j=0}^{n}\left(\begin{array}{c}n\\j\end{array}\right)\left({1\over
    a}\right)^j\cdot 1^{n-j}\\
\nonumber &=&
a^n\left(1 + \frac{1}{a}\right)^n\\
\label{eqn:identity2} \sum_{j = 0}^{n}
\left(\begin{array}{c}n\\j\end{array}\right)a^{n-j} &=& (a+1)^n
\end{eqnarray}
where $a \ge 2$ is an integer.

\begin{eqnarray}
\nonumber \sum_{j=0}^{n}\left(\begin{array}{c}n\\j\end{array}\right)j
&=&
2^{n}\sum_{j=0}^{n}\left(\begin{array}{c}n\\j\end{array}\right)j\left(\frac{1}{2}\right)^j\left(\frac{1}{2}\right)^{n-j}\\
\nonumber &=& 2^n\left[n\cdot\frac{1}{2}\right]\\
\label{eqn:identity3} \sum_{j=0}^{n}\left(\begin{array}{c}n\\j\end{array}\right)j
&=& n\cdot 2^{n-1}
\end{eqnarray}

\begin{eqnarray}
\nonumber \sum_{j=0}^{n}\left(\begin{array}{c}n\\j\end{array}\right)j^2
&=&
2^{n}\sum_{j=0}^{n}\left(\begin{array}{c}n\\j\end{array}\right)j^2\left(\frac{1}{2}\right)^j\left(\frac{1}{2}\right)^{n-j}\\
\nonumber &=& 2^n\left[\sigma^2_{\mathrm{Binomial}} +
  \mu^2_{\mathrm{Binomial}}\right]\\
\nonumber &=& 2^n\left[n\cdot\frac{1}{2}\cdot\frac{1}{2} + n^2\cdot\frac{1}{4}\right]\\
\label{eqn:identity4} \sum_{j=0}^{n}\left(\begin{array}{c}n\\j\end{array}\right)j^2
&=& n\cdot(n+1)\cdot 2^{n-2}
\end{eqnarray}

\begin{eqnarray}
\nonumber
\sum_{j=0}^{n}\left(\begin{array}{c}n\\j\end{array}\right)ja^{n-j}
\nonumber &=&(a+1)^n
\sum_{j=0}^{n}\left(\begin{array}{c}n\\j\end{array}\right)j\left(\frac{1}{a+1}\right)^j\left(\frac{a}{a+1}\right)^{n-j}\\
\nonumber &=&
(a+1)^n\left[\mu_{\mathrm{Binomial}}\left(n,\frac{1}{a+1}\right)\right]\\
\nonumber &=& (a+1)^n\left[n\cdot\frac{1}{a+1}\right] \\
\label{eqn:identity5}
\sum_{j=0}^{n}\left(\begin{array}{c}n\\j\end{array}\right)ja^{n-j} &=& n\cdot(a+1)^{n-1} 
\end{eqnarray}
Here again, $a \ge 2$ is an integer.

\begin{description}
\item[Number of expressed BBs = 0.] The number of ways either $X_1$ or
  $\bar{X}_1$ is present in $n_l - 1$ nodes is
\begin{eqnarray}
N\left(n_{{\mathrm{BB}}}^{\exp} = 0\right) &=& \sum_{j
  = 0}^{n_l-1} {\left(n_l-1\right)! \over j!\left(n_l-1-j\right)!}\\
&=& \sum_{j = 0}^{n_l-1} \left(\begin{array}{c}n_l-1\\ i\end{array}\right)\\
&=& 2^{n_l-1}
\end{eqnarray}

\item[Number of expressed BBs = 1.] Here the terminals that can be
  present in the $n_l-1$ nodes are $X_1$ or $\bar{X_1}$ or exactly one
  of the other complementary pairs. Therefore, we begin by counting
  the number of ways of having at least one copy of either $X_2$ or
  $\bar{X}_2$ in $n_l-1$ nodes. In other words, we count the number of
  ways in which only $X_2$ or its complement, $\bar{X}_2$ can be
  expressed.
\begin{eqnarray}
\nonumber &~& N\left({\mathrm{Terminals}}~{\mathrm{present}} =
  X_1~{\mathrm{or}}~\bar{X}_1~{\mathrm{or}}~X_2~{\mathrm{or}}~\bar{X}_2\right)\\
 &=& \sum_{j = 0}^{n_l-2}\sum_{k = 0}^{n_l-2-j}\sum_{q = 0}^{n_l-1-j-k} {\left(n_l-1\right)! \over j!k!q!\left(n_l-1-j-k-q\right)!}\\
&=& \sum_{j = 0}^{n_l-2} \left(\begin{array}{c}n_l-1\\
  j\end{array}\right)\sum_{k =
  0}^{n_l-2-j}\left(\begin{array}{c}n_l-1-j
  \\k\end{array}\right)\sum_{q =
  0}^{n_l-1-j-k}\left(\begin{array}{c}n_l-1-j-k\\q\end{array}\right)\\
&=& \sum_{j = 0}^{n_l-2}\left(\begin{array}{c}n_l-1\\
  j\end{array}\right)\sum_{k =
  0}^{n_l-2-j}\left(\begin{array}{c}n_l-1-j\\
  k\end{array}\right)2^{n_l-1-j-k}\\
\label{eqn:midStep0}&=& \sum_{j = 0}^{n_l-2}\left(\begin{array}{c}n_l-1\\
  j\end{array}\right)\left[\sum_{k =
  0}^{n_l-1-j}\left\{\left(\begin{array}{c}n_l-1-j\\
  k\end{array}\right)2^{n_l-1-j-k}\right\} - 1\right]\\
\label{eqn:midStep1}&=& \sum_{j = 0}^{n_l-2}\left(\begin{array}{c}n_l-1\\
  j\end{array}\right)\left[3^{n_l-1-j} - 1\right]\\
&=& \sum_{j = 0}^{n_l-1}\left(\begin{array}{c}n_l-1\\
  j\end{array}\right)\left[3^{n_l-1-j} - 1\right]\\
&=& 3^{n_l-1}\left({4 \over 3}\right)^{n_l-1} - 2^{n_l-1}\\
&=& 4^{n_l-1} - 2^{n_l-1}
\end{eqnarray}
In arriving at Equation~\ref{eqn:midStep1} from
Equation~\ref{eqn:midStep0} we use the identity given
by Equation~\ref{eqn:identity1}.\par

Note that we chose $X_2$ (or equivalently its complement, $\bar{X}_2$)
as an example. In fact there are
$\left(\begin{array}{c}m-1\\1\end{array}\right)$ alternatives to
choose from. Therefore, the total number of ways in which only one BB
gets expressed in $n_l-1$ nodes is given by
\begin{eqnarray}
N\left(n_{BB}^{\exp} = 1\right) &=& \left(\begin{array}{c} m-1\\
    1\end{array}\right)N\left({\mathrm{Terminals}}~{\mathrm{present}}
  =
  X_1~{\mathrm{or}}~\bar{X}_1~{\mathrm{or}}~X_2~{\mathrm{or}}~\bar{X}_2\right)\\
&=& (m-1)\left[4^{n_l-1} - 2^{n_l-1}\right]
\end{eqnarray}

\item[Number of expressed BBs = 2.] Here the terminals that can be
  present in the $n_l-1$ nodes are $X_1$ or $\bar{X_1}$ or exactly two
  other complementary pairs. Therefore, we begin by counting the
  number of ways of having at least one copy of either $X_2$ or
  $\bar{X}_2$ and at least one copy of either $X_3$ or $\bar{X}_3$ in
  $n_l-1$ nodes. In other words, we count the number of ways in which
  only $X_2$ or its complement, $\bar{X}_2$, and $X_3$ or its
  complement $\bar{X}_3$ can be expressed.
\begin{eqnarray}
\nonumber &~& N\left({\mathrm{Terminals}}~{\mathrm{present}} =
  X_1~{\mathrm{or}}~\bar{X}_1~{\mathrm{or}}~X_2~{\mathrm{or}}~\bar{X}_2~{\mathrm{or}}~X_3~{\mathrm{or}}~\bar{X}_3\right)\\
\nonumber  &=&
  \sum_{j=0}^{n_l-3}\sum_{k=0}^{n_l-3-j}\sum_{q=0}^{n_l-2-j-k}\sum_{r=0}^{n_l-2-j-k-q}\sum_{s=0}^{n_l-1-j-k-q-r}
  {\left(n_l-1\right)! \over j!k!q!r!s!\left(n_l-1-j-k-q-r-s\right)!}\\
&~&~  -
  \sum_{j=0}^{n_l-3}\sum_{k=0}^{n_l-3-j}\sum_{s=0}^{n_l-1-j-k}{\left(n_l-1\right)!
  \over j!k!s!\left(n_l-1-j-k-s\right)!}
\end{eqnarray}
The second summation removes the extra counting of the case when
neither $X_2$ or its complement, $\bar{X}_2$ are present in the
$n_l-1$ nodes. In other words, it ensures the presence of at least one
copy of either $X_2$ or $\bar{X}_2$.
\begin{eqnarray}
\nonumber &~& N\left({\mathrm{Terminals}}~{\mathrm{present}} =
  X_1~{\mathrm{or}}~\bar{X}_1~{\mathrm{or}}~X_2~{\mathrm{or}}~\bar{X}_2~{\mathrm{or}}~X_3~{\mathrm{or}}~\bar{X}_3\right)\\
\nonumber &=& \left[\sum_{j=0}^{n_l-3}
  \left(\begin{array}{c}n_l-1\\j\end{array}\right)\sum_{k=0}^{n_l-3-j}\left(\begin{array}{c}n_l-1-j\\k\end{array}\right)\sum_{q=0}^{n_l-2-j-k}\left(\begin{array}{c}n_l-1-j-k\\q\end{array}\right)\right.\\
\nonumber &~&~\left.\sum_{r=0}^{n_l-2-j-k-q}\left(\begin{array}{c}n_l-1-j-k-q\\r\end{array}\right)\sum_{s=0}^{n_l-1-j-k-q-r}\left(\begin{array}{c}n_l-1-j-k-q-r\\s\end{array}\right)\right]\\
&~&~  -
\label{eqn:twoBBexp}  \left[\sum_{j =
  0}^{n_l-3}\left(\begin{array}{c}n_l-1\\j\end{array}\right)\sum_{k=0}^{n_l-3-j}\left(\begin{array}{c}n_l-1-j\\k\end{array}\right)\sum_{s=0}^{n_l-1-j-k}\left(\begin{array}{c}n_l-1-j-k\\s\end{array}\right)\right]
\end{eqnarray}
Consider the sum
\[
S_2 = \left[\sum_{j =
  0}^{n_l-3}\left(\begin{array}{c}n_l-1\\j\end{array}\right)\sum_{k=0}^{n_l-3-j}\left(\begin{array}{c}n_l-1-j\\k\end{array}\right)\sum_{s=0}^{n_l-1-j-k}\left(\begin{array}{c}n_l-1-j-k\\s\end{array}\right)\right],
\]
which can be written as
\begin{eqnarray}
S_2 &=& \sum_{j = 0}^{n_l-3}\left(\begin{array}{c}n_l-1\\
  j\end{array}\right)\sum_{k =
  0}^{n_l-3-j}\left(\begin{array}{c}n_l-1-j\\
  k\end{array}\right)2^{n_l-1-j-k}\\
&=& \sum_{j = 0}^{n_l-3}\left(\begin{array}{c}n_l-1\\
  j\end{array}\right)\left[\sum_{k =
  0}^{n_l-1-j}\left\{\left(\begin{array}{c}n_l-1-j\\
  k\end{array}\right)2^{n_l-1-j-k}\right\} - 1 - 2\left(n_l-1-j\right)\right]\\
&=& \sum_{j = 0}^{n_l-3}\left(\begin{array}{c}n_l-1\\
  j\end{array}\right)\left[3^{n_l-1-j} - 1 - 2\left(n_l-1-j\right)\right]\\
&=& \sum_{j = 0}^{n_l-1}\left(\begin{array}{c}n_l-1\\
  j\end{array}\right)\left[3^{n_l-1-j} - 1 - 2\left(n_l-1-j\right)\right]\\
&=& 4^{n_l-1} - 2^{n_l-1} -
  2\left(n_l-1\right)2^{n_l-1} + 2\left(n_l-1\right)2^{n_l-2}\\
\label{eqn:s2}&=& 4^{n_l-1} - n_l2^{n_l-1}
\end{eqnarray}
The second last step in the above derivation uses the identity given
by Equation~\ref{eqn:identity3}.\par

Now consider the sum
\begin{eqnarray}
\nonumber S_1 &=& \left[\sum_{j=0}^{n_l-3}
  \left(\begin{array}{c}n_l-1\\j\end{array}\right)\sum_{k=0}^{n_l-3-j}\left(\begin{array}{c}n_l-1-j\\k\end{array}\right)\sum_{q=0}^{n_l-2-j-k}\left(\begin{array}{c}n_l-1-j-k\\q\end{array}\right)\right.\\
&~&~\left.\sum_{r=0}^{n_l-2-j-k-q}\left(\begin{array}{c}n_l-1-j-k-q\\r\end{array}\right)\sum_{s=0}^{n_l-1-j-k-q-r}\left(\begin{array}{c}n_l-1-j-k-q-r\\s\end{array}\right)\right]\\[2mm]
\nonumber &=& \left[\sum_{j=0}^{n_l-3}
  \left(\begin{array}{c}n_l-1\\j\end{array}\right)\sum_{k=0}^{n_l-3-j}\left(\begin{array}{c}n_l-1-j\\k\end{array}\right)\sum_{q=0}^{n_l-2-j-k}\left(\begin{array}{c}n_l-1-j-k\\q\end{array}\right)\right.\\
&~&~\left.\sum_{r=0}^{n_l-2-j-k-q}\left(\begin{array}{c}n_l-1-j-k-q\\r\end{array}\right)2^{n_l-1-j-k-q-r}\right]\\[2mm]
\nonumber &=& \sum_{j=0}^{n_l-3}
  \left(\begin{array}{c}n_l-1\\j\end{array}\right)\sum_{k=0}^{n_l-3-j}\left(\begin{array}{c}n_l-1-j\\k\end{array}\right)\\
&~&~\sum_{q=0}^{n_l-2-j-k}\left(\begin{array}{c}n_l-1-j-k\\q\end{array}\right)\left[3^{n_l-1-j-k-q}
  - 1\right]\\[2mm]
\nonumber &=& \sum_{j=0}^{n_l-3}
  \left(\begin{array}{c}n_l-1\\j\end{array}\right)\sum_{k=0}^{n_l-3-j}\left(\begin{array}{c}n_l-1-j\\k\end{array}\right)\left[4^{n_l-1-j-k}
  - 2^{n_l-1-j-k}\right]\\[2mm]
\nonumber &=& \sum_{j=0}^{n_l-3}
  \left(\begin{array}{c}n_l-1\\j\end{array}\right)\left[5^{n_l-1-j} -
  3^{n_l-1-j} - 2\left(n_l-1-j\right)\right]\\[2mm]
\nonumber &=& \sum_{j=0}^{n_l-1}
  \left(\begin{array}{c}n_l-1\\j\end{array}\right)\left[5^{n_l-1-j} -
  3^{n_l-1-j} - 2\left(n_l-1-j\right)\right]\\[2mm]
\nonumber &=& 6^{n_l-1} - 4^{n_l-1} - 2\left(n_l-1\right)2^{n_l-1} +
  2\left(n_l-1\right)2^{n_l-2} \\[2mm]
\label{eqn:s1} S_1 &=& 6^{n_l-1} - 4^{n_l-1} - \left(n_l-1\right)2^{n_l-1}
\end{eqnarray}

Using Equations~\ref{eqn:s1} and \ref{eqn:s2}, we get
\begin{eqnarray}
\nonumber &~& N\left({\mathrm{Terminals}}~{\mathrm{present}} =
  X_1~{\mathrm{or}}~\bar{X}_1~{\mathrm{or}}~X_2~{\mathrm{or}}~\bar{X}_2~{\mathrm{or}}~X_3~{\mathrm{or}}~\bar{X}_3\right)
  = S_1 - S_2\\
&=& \left[6^{n_l-1} - 4^{n_l-1} -
  \left(n_l-1\right)2^{n_l-1}\right]-\left[4^{n_l-1} -
  n_l2^{n_l-1}\right]\\ 
&=& 6^{n_l-1} - 2\cdot4^{n_l-1} + 2^{n_l-1}.
\end{eqnarray}

Note that we chose $X_2$ and $X_3$ (or equivalently their complement,
$\bar{X}_2$ and $\bar{X}_3$) as an example. In fact there are
$\left(\begin{array}{c}m-1\\2\end{array}\right)$ alternative pairs to
choose from. Therefore, the total number of ways in which only one BB
gets expressed in $n_l-1$ nodes is given by
\begin{eqnarray}
N\left(n_{BB}^{\exp} = 2\right) &=& \left(\begin{array}{c} m-1\\
    2\end{array}\right)N\left({\mathrm{Terminals}}~{\mathrm{present}}
  =
  X_1~{\mathrm{or}}~\bar{X}_1~{\mathrm{or}}~X_2~{\mathrm{or}}~\bar{X}_2~{\mathrm{or}}~X_3~{\mathrm{or}}~\bar{X}_3\right)\\
&=& \frac{1}{2}(m-1)(m-2)\left[6^{n_l-1} - 2\cdot4^{n_l-1} + 2^{n_l-1}\right]
\end{eqnarray}
\item[Number of expressed BBs = 3.] Here the terminals that can be
  present in the $n_l-1$ nodes are $X_1$ or $\bar{X_1}$ or exactly
  three other complementary pairs. Therefore, we begin by counting the
  number of ways of having at least one copy of either $X_2$ or
  $\bar{X}_2$, at least one copy of either $X_3$ or $\bar{X}_3$, and
  at least one copy of either $X_4$ or $\bar{X}_4$ in
  $n_l-1$ nodes. In other words, we count the number of ways in which
  only $X_2$ or its complement, $\bar{X}_2$, $X_3$ or its
  complement $\bar{X}_3$, $X_4$ or its complement $\bar{X}_4$ can be expressed.
\begin{eqnarray}
\nonumber &~& N\left({\mathrm{Terminals}}~{\mathrm{present}} =
  X_1~{\mathrm{or}}~\bar{X}_1~{\mathrm{or}}~X_2~{\mathrm{or}}~\bar{X}_2~{\mathrm{or}}~X_3~{\mathrm{or}}~\bar{X}_3~{\mathrm{or}}~X_4~{\mathrm{or}}~\bar{X}_4\right)\\
\nonumber  &=&
  \sum_{j=0}^{n_l-4}\sum_{k=0}^{n_l-4-j}\sum_{q=0}^{n_l-3-j-k}\sum_{r=0}^{n_l-3-j-k-q}\sum_{s=0}^{n_l-2-j-k-q-r}\\
\nonumber &~&~\sum_{t=0}^{n_l-2-j-k-q-r-s}\sum_{u=0}^{n_l-1-j-k-q-r-s-t}
  {\left(n_l-1\right)! \over j!k!q!r!s!t!u!\left(n_l-1-j-k-q-r-s-t-u\right)!}\\
&~&~  -
\nonumber  \sum_{j=0}^{n_l-4}\sum_{k=0}^{n_l-4-j}\sum_{q=0}^{n_l-3-j-k}\sum_{r=0}^{n_l-3-j-k-q}\sum_{u=0}^{n_l-1-j-k-q-r}{\left(n_l-1\right)!
  \over j!k!q!r!u!\left(n_l-1-j-k-q-r-u\right)!}\\
&~&~ - 
\nonumber  \sum_{j=0}^{n_l-4}\sum_{k=0}^{n_l-4-j}\sum_{s=0}^{n_l-2-j-k}\sum_{t=0}^{n_l-2-j-k-s}\sum_{u=0}^{n_l-1-j-k-s-t}{\left(n_l-1\right)!
  \over j!k!s!t!u!\left(n_l-1-j-k-s-t-u\right)!}\\
&~&~+
  \sum_{j=0}^{n_l-4}\sum_{k=0}^{n_l-4-j}\sum_{u=0}^{n_l-1-j-k}{\left(n_l-1\right)!
  \over j!k!u!\left(n_l-1-j-k-u\right)!}
\end{eqnarray}

The above equation can be rewritten as
\begin{eqnarray}
\nonumber &~& N\left({\mathrm{Terminals}}~{\mathrm{present}} =
  X_1~{\mathrm{or}}~\bar{X}_1~{\mathrm{or}}~X_2~{\mathrm{or}}~\bar{X}_2~{\mathrm{or}}~X_3~{\mathrm{or}}~\bar{X}_3~{\mathrm{or}}~X_4~{\mathrm{or}}~\bar{X}_4\right)\\
\nonumber &=& \left[\sum_{j=0}^{n_l-4}
  \left(\begin{array}{c}n_l-1\\j\end{array}\right)\sum_{k=0}^{n_l-4-j}\left(\begin{array}{c}n_l-1-j\\k\end{array}\right)\sum_{q=0}^{n_l-3-j-k}\left(\begin{array}{c}n_l-1-j-k\\q\end{array}\right)\right.\\
\nonumber
  &~&~~\sum_{r=0}^{n_l-3-j-k-q}\left(\begin{array}{c}n_l-1-j-k-q\\r\end{array}\right)\sum_{s=0}^{n_l-2-j-k-q-r}\left(\begin{array}{c}n_l-1-j-k-q-r\\s\end{array}\right)\\
\nonumber
&~&~~\sum_{t=0}^{n_l-2-j-k-q-r-s}\left(\begin{array}{c}n_l-1-j-k-q-r-s\\t\end{array}\right)\\
\nonumber
&~&~~\left.\sum_{u=0}^{n_l-1-j-k-q-r-s-t}\left(\begin{array}{c}n_l-1-j-k-q-r-s-t\\u\end{array}\right)\right]\\[2mm]
\nonumber
&~&~ - \left[\sum_{j=0}^{n_l-4}
  \left(\begin{array}{c}n_l-1\\j\end{array}\right)\sum_{k=0}^{n_l-4-j}\left(\begin{array}{c}n_l-1-j\\k\end{array}\right)\sum_{q=0}^{n_l-3-j-k}\left(\begin{array}{c}n_l-1-j-k\\q\end{array}\right)\right.\\
\nonumber
  &~&~~\left.\sum_{r=0}^{n_l-3-j-k-q}\left(\begin{array}{c}n_l-1-j-k-q\\r\end{array}\right)\sum_{u=0}^{n_l-1-j-k-q-r}\left(\begin{array}{c}n_l-1-j-k-q-r\\u\end{array}\right)\right]\\[2mm]
\nonumber &~&~ - \left[\sum_{j=0}^{n_l-4}
  \left(\begin{array}{c}n_l-1\\j\end{array}\right)\sum_{k=0}^{n_l-4-j}\left(\begin{array}{c}n_l-1-j\\k\end{array}\right)\sum_{s=0}^{n_l-2-j-k}\left(\begin{array}{c}n_l-1-j-k\\s\end{array}\right)\right.\\
\nonumber
&~&~~\left.\sum_{t=0}^{n_l-2-j-k-s}\left(\begin{array}{c}n_l-1-j-k-s\\t\end{array}\right)\sum_{u=0}^{n_l-1-j-k-s-t}\left(\begin{array}{c}n_l-1-j-k-s-t\\u\end{array}\right)\right]\\[2mm]
&~&~ + 
\left[\sum_{j = 0}^{n_l-4}\left(\begin{array}{c}n_l-1\\j\end{array}\right)\sum_{k=0}^{n_l-4-j}\left(\begin{array}{c}n_l-1-j\\k\end{array}\right)\sum_{u=0}^{n_l-1-j-k}\left(\begin{array}{c}n_l-1-j-k\\u\end{array}\right)\right]
\end{eqnarray}
Consider the sum
\[
S_4 = \left[\sum_{j =
  0}^{n_l-4}\left(\begin{array}{c}n_l-1\\j\end{array}\right)\sum_{k=0}^{n_l-4-j}\left(\begin{array}{c}n_l-1-j\\k\end{array}\right)\sum_{u=0}^{n_l-1-j-k}\left(\begin{array}{c}n_l-1-j-k\\u\end{array}\right)\right],
\]
which can be written as
\begin{eqnarray}
S_4 &=& \sum_{j = 0}^{n_l-4}\left(\begin{array}{c}n_l-1\\
  j\end{array}\right)\sum_{k =
  0}^{n_l-4-j}\left(\begin{array}{c}n_l-1-j\\
  k\end{array}\right)2^{n_l-1-j-k}\\
\nonumber &=& \sum_{j = 0}^{n_l-4}\left(\begin{array}{c}n_l-1\\
  j\end{array}\right)\left[\sum_{k =
  0}^{n_l-1-j}\left\{\left(\begin{array}{c}n_l-1-j\\
  k\end{array}\right)2^{n_l-1-j-k}\right\} - 1\right.\\
&~&~~ \left.- 2\left(n_l-1-j\right) - 2\left(n_l-1-j\right)\left(n_l-2-j\right)\right]\\
&=& \sum_{j = 0}^{n_l-4}\left(\begin{array}{c}n_l-1\\
  j\end{array}\right)\left[3^{n_l-1-j} - 1 - 2\left(n_l-1-j\right) - 2\left(n_l-1-j\right)\left(n_l-2-j\right)\right]\\
&=& \sum_{j = 0}^{n_l-1}\left(\begin{array}{c}n_l-1\\
  j\end{array}\right)\left[3^{n_l-1-j} - 1 - 2\left(n_l-1-j\right) - 2\left(n_l-1-j\right)\left(n_l-2-j\right)\right]\\
&=& \sum_{j = 0}^{n_l-1}\left(\begin{array}{c}n_l-1\\
  j\end{array}\right)\left[3^{n_l-1-j} - 
\left(2\left(n_l-1\right)\left(n_l-2\right) + 2n_l - 1\right) + 
4\left(n_l-1\right)j - 2j^2\right]\\
&=& 4^{n_l-1} - \left[2\left(n_l-1\right)\left(n_l-2\right) + 2n_l -
  1\right]2^{n_l-1} + 4\left(n_l-1\right)^22^{n_l-2} - 2n_l\left(n_l-1\right)2^{n_1-3}\\
\label{eqn:s4a}&=& 4^{n_l-1} - 2^{n_l-1} - n_l\left(n_l-1\right)2^{n_l-2}
\end{eqnarray}

\begin{eqnarray}
\nonumber S_3 &=& \left[\sum_{j=0}^{n_l-4}
  \left(\begin{array}{c}n_l-1\\j\end{array}\right)\sum_{k=0}^{n_l-4-j}\left(\begin{array}{c}n_l-1-j\\k\end{array}\right)\sum_{s=0}^{n_l-2-j-k}\left(\begin{array}{c}n_l-1-j-k\\q\end{array}\right)\right.\\
&~&~\left.\sum_{t=0}^{n_l-2-j-k-s}\left(\begin{array}{c}n_l-1-j-k-s\\t\end{array}\right)\sum_{u=0}^{n_l-1-j-k-s-t}\left(\begin{array}{c}n_l-1-j-k-s-t\\u\end{array}\right)\right]\\[2mm]
\nonumber &=& \left[\sum_{j=0}^{n_l-4}
  \left(\begin{array}{c}n_l-1\\j\end{array}\right)\sum_{k=0}^{n_l-4-j}\left(\begin{array}{c}n_l-1-j\\k\end{array}\right)\sum_{s=0}^{n_l-2-j-k}\left(\begin{array}{c}n_l-1-j-k\\s\end{array}\right)\right.\\
&~&~\left.\sum_{t=0}^{n_l-2-j-k-s}\left(\begin{array}{c}n_l-1-j-k-s\\t\end{array}\right)2^{n_l-1-j-k-s-t}\right]\\[2mm]
\nonumber &=& \sum_{j=0}^{n_l-4}
  \left(\begin{array}{c}n_l-1\\j\end{array}\right)\sum_{k=0}^{n_l-4-j}\left(\begin{array}{c}n_l-1-j\\k\end{array}\right)\\
&~&~\sum_{s=0}^{n_l-2-j-k}\left(\begin{array}{c}n_l-1-j-k\\s\end{array}\right)\left[3^{n_l-1-j-k-s}
  - 1\right]\\[2mm]
&=& \sum_{j=0}^{n_l-4}
  \left(\begin{array}{c}n_l-1\\j\end{array}\right)\sum_{k=0}^{n_l-4-j}\left(\begin{array}{c}n_l-1-j\\k\end{array}\right)\left[4^{n_l-1-j-k}
  - 2^{n_l-1-j-k}\right]\\[2mm]
&=& \sum_{j=0}^{n_l-4}
  \left(\begin{array}{c}n_l-1\\j\end{array}\right)\left[5^{n_l-1-j} -
  3^{n_l-1-j} - 2\left(n_l-1-j\right) - 6\left(n_l-1-j\right)\left(n_l-2-j\right)\right]\\[2mm]
 &=& \sum_{j=0}^{n_l-1}
  \left(\begin{array}{c}n_l-1\\j\end{array}\right)\left[5^{n_l-1-j} -
  3^{n_l-1-j} - 2\left(n_l-1-j\right) -
  6\left(n_l-1-j\right)\left(n_l-2-j\right)\right]\\[2mm]
 &=& 6^{n_l-1} - 4^{n_l-1} - \sum_{j=0}^{n_l-1}
  \left(\begin{array}{c}n_l-1\\j\end{array}\right)\left[6\left(n_l-1\right)^2
  - 4\left(n_l-1\right) - 12\left(n_l-1\right)j + 4j +
  6j^2\right]\\[2mm]
\nonumber &=& 6^{n_l-1} - 4^{n_l-1} - 6\left(n_l-1\right)^22^{n_l-1} +
  4\left(n_l-1\right)2^{n_l-1} + 12\left(n_l-1\right)^22^{n_l-2}\\
&~&~~ -
  4\left(n_l-1\right)2^{n_l-2} - 6n_l\left(n_l-1\right)2^{n_l-3}\\[2mm]
\label{eqn:s3a} S_3 &=& 6^{n_l-1} - 4^{n_l-1} +
  2\left(n_l-1\right)2^{n_l-1} - 3n_l\left(n_l-1\right)2^{n_l-2}
\end{eqnarray}

\begin{eqnarray}
\nonumber S_2 &=& \left[\sum_{j=0}^{n_l-4}
  \left(\begin{array}{c}n_l-1\\j\end{array}\right)\sum_{k=0}^{n_l-4-j}\left(\begin{array}{c}n_l-1-j\\k\end{array}\right)\sum_{q=0}^{n_l-3-j-k}\left(\begin{array}{c}n_l-1-j-k\\q\end{array}\right)\right.\\
&~&~\left.\sum_{r=0}^{n_l-3-j-k-q}\left(\begin{array}{c}n_l-1-j-k-q\\r\end{array}\right)\sum_{u=0}^{n_l-1-j-k-q-r}\left(\begin{array}{c}n_l-1-j-k-q-r\\u\end{array}\right)\right]\\[2mm]
\nonumber &=& \left[\sum_{j=0}^{n_l-4}
  \left(\begin{array}{c}n_l-1\\j\end{array}\right)\sum_{k=0}^{n_l-4-j}\left(\begin{array}{c}n_l-1-j\\k\end{array}\right)\sum_{q=0}^{n_l-3-j-k}\left(\begin{array}{c}n_l-1-j-k\\q\end{array}\right)\right.\\
&~&~\left.\sum_{r=0}^{n_l-3-j-k-q}\left(\begin{array}{c}n_l-1-j-k-q\\r\end{array}\right)2^{n_l-1-j-k-q-r}\right]\\[2mm]
\nonumber &=& \sum_{j=0}^{n_l-4}
  \left(\begin{array}{c}n_l-1\\j\end{array}\right)\sum_{k=0}^{n_l-4-j}\left(\begin{array}{c}n_l-1-j\\k\end{array}\right)\\
&~&~\sum_{q=0}^{n_l-3-j-k}\left(\begin{array}{c}n_l-1-j-k\\s\end{array}\right)\left[3^{n_l-1-j-k-q}
  - 1 - 2\left(n_l-1-j-k-q\right)\right]\\[2mm]
\nonumber &=& \sum_{j=0}^{n_l-4}
  \left(\begin{array}{c}n_l-1\\j\end{array}\right)\sum_{k=0}^{n_l-4-j}\left(\begin{array}{c}n_l-1-j\\k\end{array}\right)\left[4^{n_l-1-j-k}
  - \left(2n_l-1-2j-2k\right)2^{n_l-1-j-k}\right.\\
&~&~~\left. + 2\left(n_l-1-j-k\right)2^{n_l-2-j-k}\right]\\[2mm]
&=& \sum_{j=0}^{n_l-4}
  \left(\begin{array}{c}n_l-1\\j\end{array}\right)\sum_{k=0}^{n_l-4-j}\left(\begin{array}{c}n_l-1-j\\k\end{array}\right)\left[4^{n_l-1-j-k}
  - \left(n_l-j-k\right)2^{n_l-1-j-k}\right]\\[2mm]
\nonumber &=& \sum_{j=0}^{n_l-4}
  \left(\begin{array}{c}n_l-1\\j\end{array}\right)\left\{\sum_{k=0}^{n_l-1-j}\left(\begin{array}{c}n_l-1-j\\k\end{array}\right)\left[4^{n_l-1-j-k}
  - \left(n_l-j-k\right)2^{n_l-1-j-k}\right]\right.\\
&~&~~\left. - 2\left(n_l-1-j\right)\left(n_l-2-j\right)\right\}\\[2mm]
\nonumber &=& \sum_{j=0}^{n_l-4}
  \left(\begin{array}{c}n_l-1\\j\end{array}\right)\left[5^{n_l-1-j} -
  \left(n_l-j\right)3^{n_l-1-j} +
  \left(n_l-1-j\right)3^{n_l-2-j}\right.\\
&~&~~\left. - 2\left(n_l-1-j\right)\left(n_l-2-j\right)\right]\\[2mm]
\nonumber &=& \sum_{j=0}^{n_l-1}
  \left(\begin{array}{c}n_l-1\\j\end{array}\right)\left[5^{n_l-1-j} -
  \left(n_l-j\right)3^{n_l-1-j} +
  \left(n_l-1-j\right)3^{n_l-2-j}\right.\\
&~&~~\left. - 2\left(n_l-1-j\right)\left(n_l-2-j\right)\right]\\[2mm]
\nonumber &=& 6^{n_l-1} - n_l4^{n_1-1} +
  \frac{1}{3}\left(n_l-1\right)4^{n_l-1} -
  2\left(n_l-1\right)\left(n_l-2\right)2^{n_l-1} +\\
&~&~~ 2\left(2n_l-3\right)\sum_{j=0}^{n_l-1}
  \left(\begin{array}{c}n_l-1\\j\end{array}\right)j - 2\sum_{j=0}^{n_l-1}
  \left(\begin{array}{c}n_l-1\\j\end{array}\right)j^2 + \frac{2}{3} \sum_{j=0}^{n_l-1}
  \left(\begin{array}{c}n_l-1\\j\end{array}\right)j3^{n_l-1-j}\\[2mm]
\nonumber &=& 6^{n_l-1} - n_l4^{n_1-1} +
  \frac{1}{3}\left(n_l-1\right)4^{n_l-1} -
  2\left(n_l-1\right)\left(n_l-2\right)2^{n_l-1} +\\
&~&~~ 2\left(2n_l-3\right)\left(n_l-1\right)2^{n_l-2} -
  2n_l\left(n_l-1\right)2^{n_l-3} +
  \frac{2}{3}\left(n_l-1\right)4^{n_l-2}\\[2mm]
\label{eqn:s2a} S_2 &=& 6^{n_l-1} - \frac{1}{2}\left(n_l+1\right)4^{n_1-1} +
  \left(n_l-1\right)2^{n_l-1} - n_l\left(n_l-1\right)2^{n_l-2}
\end{eqnarray}

\begin{eqnarray}
\nonumber S_1 &=& \sum_{j=0}^{n_l-4}
  \left(\begin{array}{c}n_l-1\\j\end{array}\right)\sum_{k=0}^{n_l-4-j}\left(\begin{array}{c}n_l-1-j\\k\end{array}\right)\sum_{q=0}^{n_l-3-j-k}\left(\begin{array}{c}n_l-1-j-k\\q\end{array}\right)\\
\nonumber
  &~&~~\sum_{r=0}^{n_l-3-j-k-q}\left(\begin{array}{c}n_l-1-j-k-q\\r\end{array}\right)\sum_{s=0}^{n_l-2-j-k-q-r}\left(\begin{array}{c}n_l-1-j-k-q-r\\s\end{array}\right)\\
\nonumber
&~&~~\sum_{t=0}^{n_l-2-j-k-q-r-s}\left(\begin{array}{c}n_l-1-j-k-q-r-s\\t\end{array}\right)\\
&~&~~\sum_{u=0}^{n_l-1-j-k-q-r-s-t}\left(\begin{array}{c}n_l-1-j-k-q-r-s-t\\u\end{array}\right)\\[2mm]
\nonumber &=& \sum_{j=0}^{n_l-4}
  \left(\begin{array}{c}n_l-1\\j\end{array}\right)\sum_{k=0}^{n_l-4-j}\left(\begin{array}{c}n_l-1-j\\k\end{array}\right)\sum_{q=0}^{n_l-3-j-k}\left(\begin{array}{c}n_l-1-j-k\\q\end{array}\right)\\
\nonumber
  &~&~~\sum_{r=0}^{n_l-3-j-k-q}\left(\begin{array}{c}n_l-1-j-k-q\\r\end{array}\right)\sum_{s=0}^{n_l-2-j-k-q-r}\left(\begin{array}{c}n_l-1-j-k-q-r\\s\end{array}\right)\\
&~&~~\sum_{t=0}^{n_l-2-j-k-q-r-s}\left(\begin{array}{c}n_l-1-j-k-q-r-s\\t\end{array}\right)
  2^{n_l-1-j-k-q-r-s-t}\\[2mm]
\nonumber &=& \sum_{j=0}^{n_l-4}
  \left(\begin{array}{c}n_l-1\\j\end{array}\right)\sum_{k=0}^{n_l-4-j}\left(\begin{array}{c}n_l-1-j\\k\end{array}\right)\sum_{q=0}^{n_l-3-j-k}\left(\begin{array}{c}n_l-1-j-k\\q\end{array}\right)\\
\nonumber
  &~&~~\sum_{r=0}^{n_l-3-j-k-q}\left(\begin{array}{c}n_l-1-j-k-q\\r\end{array}\right)\sum_{s=0}^{n_l-2-j-k-q-r}\left(\begin{array}{c}n_l-1-j-k-q-r\\s\end{array}\right)\\
&~&~~\left\{3^{n_l-1-j-k-q-r-s} - 1\right\}\\[2mm]
\nonumber &=& \sum_{j=0}^{n_l-4}
  \left(\begin{array}{c}n_l-1\\j\end{array}\right)\sum_{k=0}^{n_l-4-j}\left(\begin{array}{c}n_l-1-j\\k\end{array}\right)\sum_{q=0}^{n_l-3-j-k}\left(\begin{array}{c}n_l-1-j-k\\q\end{array}\right)\\
\nonumber
  &~&~~\sum_{r=0}^{n_l-3-j-k-q}\left(\begin{array}{c}n_l-1-j-k-q\\r\end{array}\right)\sum_{s=0}^{n_l-1-j-k-q-r}\left(\begin{array}{c}n_l-1-j-k-q-r\\s\end{array}\right)\\
&~&~~\left\{3^{n_l-1-j-k-q-r-s} - 1\right\}\\[2mm]
\nonumber &=& \sum_{j=0}^{n_l-4}
  \left(\begin{array}{c}n_l-1\\j\end{array}\right)\sum_{k=0}^{n_l-4-j}\left(\begin{array}{c}n_l-1-j\\k\end{array}\right)\sum_{q=0}^{n_l-3-j-k}\left(\begin{array}{c}n_l-1-j-k\\q\end{array}\right)\\
  &~&~~\sum_{r=0}^{n_l-3-j-k-q}\left(\begin{array}{c}n_l-1-j-k-q\\r\end{array}\right)\left\{4^{n_l-1-j-k-q-r}
  - 2^{n_l-1-j-k-q-r}\right\}\\[2mm]
\nonumber &=& \sum_{j=0}^{n_l-4}
  \left(\begin{array}{c}n_l-1\\j\end{array}\right)\sum_{k=0}^{n_l-4-j}\left(\begin{array}{c}n_l-1-j\\k\end{array}\right)\sum_{q=0}^{n_l-3-j-k}\left[\left(\begin{array}{c}n_l-1-j-k\\q\end{array}\right)\right.\\
\nonumber &~&~~\sum_{r=0}^{n_l-1-j-k-q}\left(\begin{array}{c}n_l-1-j-k-q\\r\end{array}\right)\left\{4^{n_l-1-j-k-q-r}
  - 2^{n_l-1-j-k-q-r}\right\}\\
&~&\left. - 2\left(n_l-1-j-k-q\right)\right]\\[2mm]
\nonumber &=& \sum_{j=0}^{n_l-4}
  \left(\begin{array}{c}n_l-1\\j\end{array}\right)\sum_{k=0}^{n_l-4-j}\left(\begin{array}{c}n_l-1-j\\k\end{array}\right)\sum_{q=0}^{n_l-3-j-k}\left(\begin{array}{c}n_l-1-j-k\\q\end{array}\right)\\
&~&\left\{5^{n_l-1-j-k-q} - 3^{n_l-1-j-k-q} - 2\left(n_l-1-j-k-q\right)\right\}\\[2mm]
\nonumber &=& \sum_{j=0}^{n_l-4}
  \left(\begin{array}{c}n_l-1\\j\end{array}\right)\sum_{k=0}^{n_l-4-j}\left(\begin{array}{c}n_l-1-j\\k\end{array}\right)\sum_{q=0}^{n_l-1-j-k}\left(\begin{array}{c}n_l-1-j-k\\q\end{array}\right)\\
&~&\left\{5^{n_l-1-j-k-q} - 3^{n_l-1-j-k-q} -
  2\left(n_l-1-j-k-q\right)\right\}\\[2mm]
\nonumber &=& \sum_{j=0}^{n_l-4}
  \left(\begin{array}{c}n_l-1\\j\end{array}\right)\sum_{k=0}^{n_l-4-j}\left(\begin{array}{c}n_l-1-j\\k\end{array}\right)\left\{6^{n_l-1-j-k}
  - 4^{n_l-1-j-k}\right.\\
&~&~~\left. -2\left(n_l-1-j-k\right)2^{n_l-1-j-k} + 2\left(n_l-1-j-k\right)2^{n_l-2-j-k}\right\}\\[2mm]
\nonumber &=& \sum_{j=0}^{n_l-4}
  \left(\begin{array}{c}n_l-1\\j\end{array}\right)\sum_{k=0}^{n_l-4-j}\left(\begin{array}{c}n_l-1-j\\k\end{array}\right)\left\{6^{n_l-1-j-k}
  - 4^{n_l-1-j-k}\right.\\
&~&~~\left.- \left(n_l-1-j-k\right)2^{n_l-1-j-k}\right\}\\[2mm]
\nonumber &=& \sum_{j=0}^{n_l-4}\left[
  \left(\begin{array}{c}n_l-1\\j\end{array}\right)\sum_{k=0}^{n_l-1-j}\left(\begin{array}{c}n_l-1-j\\k\end{array}\right)\left\{6^{n_l-1-j-k}
  - 4^{n_l-1-j-k}\right.\right.\\
&~&~~\left.\left.- \left(n_l-1-j-k\right)2^{n_l-1-j-k}\right\} - 6\left(n_l-1-j\right)\left(n_l-2-j\right)\right]\\[2mm]
\nonumber &=& \sum_{j=0}^{n_l-4}
  \left(\begin{array}{c}n_l-1\\j\end{array}\right)\left\{7^{n_l-1-j} -
  5^{n_l-1-j} - \left(n_l-1-j\right)3^{n_l-1-j} \right.\\
&~&~~+ \left.\left(n_l-1-j\right)3^{n_l-2-j} - 6\left(n_l-1-j\right)\left(n_l-2-j\right)\right\}\\[2mm]
\nonumber &=& \sum_{j=0}^{n_l-4}\left(\begin{array}{c}n_l-1\\j\end{array}\right)\left\{7^{n_l-1-j} -
  5^{n_l-1-j} - \frac{2}{3}\left(n_l-1-j\right)3^{n_l-1-j} \right.\\
&~&~~\left.-6\left(n_l-1-j\right)\left(n_l-2-j\right)\right\}\\[2mm]
\nonumber &=& \sum_{j=0}^{n_l-1}\left(\begin{array}{c}n_l-1\\j\end{array}\right)\left\{7^{n_l-1-j} -
  5^{n_l-1-j} - \frac{2}{3}\left(n_l-1-j\right)3^{n_l-1-j} \right.\\
&~&~~\left. - 6\left(n_l-1-j\right)\left(n_l-2-j\right)\right\}\\[2mm]
\nonumber &=& 8^{n_l-1-j} - 6^{n_l-1-j} -
  \frac{2}{3}\left(n_l-1\right)4^{n_l-1} +
  \frac{2}{3}\left(n_l-1\right)4^{n_l-2} -
  6\left(n_l-1\right)\left(n_l-2\right)2^{n_l-1}\\
&~&~~ +
  6\left(2n_l-3\right)\left(n_l-1\right)2^{n_l-2} -
  6n_l\left(n_l-1\right)2^{n_l-3}\\[2mm]
&=& 8^{n_l-1-j} - 6^{n_l-1-j} -
  \frac{1}{2}\left(n_l-1\right)4^{n_l-1} +
  3\left(n_l-1\right)2^{n_l-1} - 3n_l\left(n_l-1\right)2^{n_l-2}\\[2mm]
\label{eqn:s1a} S_1 &=& 8^{n_l-1} - 6^{n_l-1} + 4^{n_l-1} -
  \frac{1}{2}\left(n_l+1\right)4^{n_l-1} + 3\left(n_l-1\right)2^{n_l-1}  - 3n_l\left(n_l-1\right)2^{n_l-2}
\end{eqnarray}
Using Equations~\ref{eqn:s4a} -- \ref{eqn:s1a}, we get
\begin{eqnarray}
\nonumber && N\left({\mathrm{Terminals}}~{\mathrm{present}} =
  X_1~{\mathrm{or}}~\bar{X}_1~{\mathrm{or}}~X_2~{\mathrm{or}}~\bar{X}_2~{\mathrm{or}}~X_3~{\mathrm{or}}~\bar{X}_3~{\mathrm{or}}~X_4~{\mathrm{or}}~\bar{X}_4\right)
  = S_1 - S_2 - S_3 + S4\\[2mm]
\nonumber &=& \left[8^{n_l-1} - 6^{n_l-1} + 4^{n_l-1} -
  \frac{1}{2}\left(n_l+1\right)4^{n_l-1} +
  3\left(n_l-1\right)2^{n_l-1} -
  3n_l\left(n_l-1\right)2^{n_l-2}\right]\\
\nonumber &~&~~-\left[6^{n_l-1} - \frac{1}{2}\left(n_l+1\right)4^{n_1-1} +
  \left(n_l-1\right)2^{n_l-1} -
  n_l\left(n_l-1\right)2^{n_l-2}\right]\\
\nonumber &~&~~-\left[6^{n_l-1} - 4^{n_l-1} +
  2\left(n_l-1\right)2^{n_l-1} - 3n_l\left(n_l-1\right)2^{n_l-2}\right]\\
&~&~~+\left[4^{n_l-1} - 2^{n_l-1} - n_l\left(n_l-1\right)2^{n_l-2}\right]\\[2mm]
&=& 8^{n_l-1} - 3\cdot6^{n_l-1} + 3\cdot4^{n_l-1} - 2\cdot2^{n_l-1}.
\end{eqnarray}

Note that we chose $X_2$, $X_3$, and $X_4$ (or equivalently their complement,
$\bar{X}_2$, $\bar{X}_3$, and $\bar{X}_4$) as an example. In fact there are
$\left(\begin{array}{c}m-1\\3\end{array}\right)$ alternative pairs to
choose from. Therefore, the total number of ways in which only one BB
gets expressed in $n_l-1$ nodes is given by
\begin{eqnarray}
N\left(n_{BB}^{\exp} = 3\right) &=& \left(\begin{array}{c} m-1\\
    3\end{array}\right)N\left({\mathrm{Terminals}}~=
  X_1~{\mathrm{or}}~\bar{X}_1~{\mathrm{or}}~X_2~{\mathrm{or}}~\bar{X}_2~{\mathrm{or}}~X_3~{\mathrm{or}}~\bar{X}_3~{\mathrm{or}}~X_4~{\mathrm{or}}~\bar{X}_4\right)\\
&=& \left(\begin{array}{c} m-1\\
    3\end{array}\right)\left[8^{n_l-1} - 3\cdot6^{n_l-1} + 3\cdot4^{n_l-1} - 2^{n_l-1}\right]
\end{eqnarray}
\end{description}
From the above cases we can generalize the number of ways of
expressing $i$ BBs in $n_l-1$ nodes is given by
\begin{equation}
N\left(n_{BB}^{\exp} = i\right) = \left(\begin{array}{c} m-1\\
    i\end{array}\right)\sum_{j = 0}^{i}\left(\begin{array}{c} i\\
    j\end{array}\right)\left(-1\right)^j\left[2\left(i-j+1\right)\right]^{n_l-1}
\end{equation}

Recall that the total number of ways of arranging the $2m$ terminals
in $n_l-1$ nodes is given by
\[
N_{{\mathrm{tot}}} = \left(2m\right)^{n_l-1}
\] 

Therefore, the probability of expressing $i$ BBs is given by
\begin{eqnarray}
p\left(n_{BB}^{\exp} = i\right) &=& {N\left(n_{BB}^{\exp} = i\right)
  \over N_{{\mathrm{tot}}}}\\
&=& \left(\begin{array}{c} m-1\\
    i\end{array}\right)\sum_{j = 0}^{i}\left(\begin{array}{c} i\\
    j\end{array}\right)\left(-1\right)^j\left({i-j+1 \over m}\right)^{n_l-1}
\end{eqnarray}

The average number of expressed building blocks other than the one
that decision is being made on
\begin{equation}
\bar{n}_{BB}^{\exp} = \sum_{i = 0}^{m-1}\left(\begin{array}{c} m-1\\
    i\end{array}\right)i\sum_{j = 0}^{i}\left(\begin{array}{c} i\\
    j\end{array}\right)\left(-1\right)^j\left({i-j+1 \over m}\right)^{n_l-1} 
\end{equation}

The variance in the number of expressed building blocks other than the one
that decision is being made on
\begin{equation}
\sigma^2_{{n}_{BB}^{\exp}} = \sum_{i = 0}^{m-1}\left(\begin{array}{c} m-1\\
    i\end{array}\right)i^2\sum_{j = 0}^{i}\left(\begin{array}{c} i\\
    j\end{array}\right)\left(-1\right)^j\left({i-j+1 \over
    m}\right)^{n_l-1}  - \left[\bar{n}_{BB}^{\exp}\right]^2
\end{equation}

\section{Estimating Tree Sizes}
We start with defining two utility procedures that generate a non-full
tree and full tree respectively. We have named them accordingly and
they correspond in common GP parlance to GROW and FULL.  These
procedures are called by RAMPED-FULL, RAMPED-GROW and
RAMPED-HALF-HALF.

Both algorithms are parameterized by:
\begin{itemize}
\item $maxHeight$ : the maximum allowable height of the tree
\item $q$: the probability with which the terminal set is used to draw a new tree node
\end{itemize}
Often $q$ is implicitly set as the frequency of terminal nodes in the
primitive set and GPr's simply set maxHeight. However, sometimes (like
we do in the ORDER problem) a bias between functions and terminals is
introduced.  We note that Luke \cite{luke:2000:2ftcaGP} has similar versions of
these algorithms without $q$ explicitized.

\begin{listing}{1}
Algorithm I: create-tree-not-necessarily-full (q, maxHeight)
// create trees of more than 1 node
root = get-function()
height = 1
left-child = create-subtree(q, maxHeight-1, height)
right-child = create-subtree(q, maxHeight-1,height)
return make-tree(root, left-child,right-child)

procedure create-subtree(q, maxHeight, current-height)
if current-height = (maxHeight-1)
then
   return get-terminal()
else
   if rand(0,1) < q then
       return get-terminal()
   else
       return create-tree-not-necessarily-full(q, maxHeight-1)
\end{listing}

The {\it create-tree-not-necessarily-full} algorithm creates a GP tree of
height between 2 and maxHeight, not allowing a single leaf to be
generated as a tree.  The tree is not necessarily full. After drawing
a function for the tree's root node, it uses $q$ to decide between
making each child subtree of the root a function or a terminal, {\it
except} when the tree's height is equal to $(maxHeight -1)$. When the
tree's height is equal to $maxHeight -1$, a terminal is alway
generated as the child subtree.  This ensures that no tree has height
greater than $maxHeight$.

\begin{listing}{1}
Algorithm II: create-tree-full (q, maxHeight)
// create full trees of more than 1 node
root = get-function()
height = 1
left-child = create-full-subtree(q, maxHeight-1, height)
right-child = create-tree-full(q, height(left-child))
return make-tree(root, left-child,right-child)

procedure create-full-subtree(q, maxHeight, current-height)
if current-height = (maxHeight-1)
then
   return get-terminal()
else
   if rand(0,1) < q then
       return get-terminal()
   else
       return create-tree-full(q, maxHeight-1)
\end{listing}

The {\it create-tree-full} algorithm creates a GP tree of
height between 2 and maxHeight, not allowing a single leaf to be
generated as a tree.  The tree is always full. After drawing
a function for the tree's root node, it uses $q$ to decide between
making the left child subtree of the root a function or a terminal, {\it
except} when the tree's height is equal to $(maxHeight -1)$. When the
tree's height is equal to $maxHeight -1$, a terminal is alway
generated as the child subtree.  This ensures that no tree has height
greater than $maxHeight$.  The right child subtree of the root is
generated by calling {\it create-tree-full} with the maxHeight
parameter taking the value of the height of the left child
subtree. (UM: But I haven't checked my pseudocode carefully at all)

Usually these procedures are subsumed by procedures that create an
initial population with random {\it fitness} but predetermined expected
GP tree structure. The procedures are:
\begin{itemize}
\item ramped full. Create subsamples of trees for each height, h,  between
  height 1 and maxHeight. Each subsample has full trees of height up
  to h.
\item ramped not-necessarily-full. Create subsamples of trees for each
  height, h, between 1 and maxHeight. Each subsample has
  not-necessarily full trees of height up to h. 
\item ramped half-half (implying half full and half not necessarily
  full). Create two subsamples for each height, h, between height 1
  and maxHeight. One subsample has full trees of height up to h and
  one subsample has not-necessarily full trees of height up to h.
\end{itemize}

Assuming any of these algorithms is executed to generate a tree of size $s$,
because the tree is binary, the following is known,:
\begin{enumerate}
\item the number of leaves (terminals), $t_s = \frac{s+1}{2}$
\item the number of internal nodes (functions), $f_s = \frac{s-1}{2}$

\end{enumerate}

The average size of a tree created using Algorithm {\it
  create-tree-not-necessarily-full} can be estimated as follows:

\begin{enumerate}
\item a tree of size h has a range of possible sizes from
$s_{\min} = 2h+1$ to 
$s_{\max} = 2^{h+1}-1$. This range is
$s_{\min}, s_{\min}+2, \dots,s_{\max}$.
\item the probability of a tree of size $s$ given it has height $h$
 and $h < h_{\max}$, where $h_{\max}$ is maxHeight:
\begin{equation}
p\left(s|h;h<h_{\max}\right) = (1-q)^{f_s-1}q^{t_s}
\end{equation}
\item the probability of a tree of $h < h_{\max}$:
\begin{equation}
p\left(h<h_{\max}\right) = \sum_{h=1}^{h_{\max}-1}\sum_{s=s_{\min} by 2}^{s=s_{\max}} p(h|s)
\end{equation}
\item the average size of a tree of height h:
\begin{equation}
\overline{s}(h) = 
\sum_{s=s_{\min} by 2}^{s=s_{\max}} p(s|h)s
\end{equation}
\item the average size of trees of height $h < h_{\max}$ 
\begin{equation}
\overline{s}(h;h<h_{\max}) = \sum_{h=1}^{h_{\max}-1} \overline{s}(h)\|p(h|s)\|
\end{equation}
\item the estimated average size of a tree of height, $h = h_{\max}$ can be 
estimated conservatively (underestimation):
\begin{equation}
\hat{s}\left(h = h_{\max}\right) =
\end{equation}
\item the probability of a tree of height = $maxHeight$:
\begin{equation}
p\left(h = h_{\max}\right) = 1 - p\left(h < h_{\max}\right)
\end{equation}
\item the estimated average size of any tree:
\begin{equation}
\hat{s} = p\left(h = h_{\max}\right)\hat{s}\left(h = h_{\max}\right) +
               p\left(h < h_{\max}\right)\overline{s}\left(h < h_{\max}\right)
\end{equation}
\end{enumerate}

The average size of a tree created using Algorithm {\it
  create-tree-full} can be estimated as follows:

\begin{enumerate}
%%\item the probability of a tree of height=1:  $p(1) = q$
%%\item the probability of a tree of height=2:  $p(2) = (1-q)q$
\item the probability of a tree of height h when $h < maxHeight$:
\begin{equation} 
p(h) = (1-q)^{h-1}q
\end{equation}
\item the probability of any tree of height, $h$, that is less than
  the maxHeight, $h_{\max}$:
\begin{eqnarray} 
\nonumber p\left(h<h_{\max}\right) &=& \sum_{h=1}^{h_{\max}-1}p(h),\\
\nonumber &=& \sum_{h=1}^{h_{\max}-1}q(1-q)^{h-1},\\
\nonumber &=& 1 - (1-q)^{h_{\max}-1}.
\end{eqnarray}
\item the probability of a tree of $h = h_{\max}$:
\begin{eqnarray}
\nonumber p\left(h=h_{\max}\right) &=& 1 - p\left(h<h_{\max}\right),\\
&=& (1-q)^{h_{\max}-1}.
\end{eqnarray}
\item the size of a tree of height h, $s(h) = 2^{h+1}-1$
\item the average size of a tree of height, $h < h_{\max}$:
\begin{eqnarray}
\nonumber \overline{s}\left(h < h_{\max}\right) &=&
\sum_{h=1}^{h_{\max}-1}\|p(h)\|s(h),\\
\nonumber &=& \left[{1 \over 1 - (1-q)^{h_{\max}-1}}\right]\sum_{h=1}^{h_{\max}-1}
\left[\left(2^{h+1}-1\right)q(1-q)^(h-1)\right],\\
 &=& \left({4q \over 2q-1}\right)\left[{1 -
  \left(2(1-q)\right)^{h_{\max}-1} \over 1 - (1-q)^{h_{\max}-1}}\right] - 1.
\end{eqnarray}
\item the average size of a tree of height, $h = h_{\max}$,
\begin{eqnarray}
\nonumber \overline{s}\left(h = h_{\max}\right) &=&
\|p\left(h=h_{\max}\right)\|s\left(h = h_{\max}\right), \\
&=& 2^{h_{\max}+1}-1.
\end{eqnarray}
\item the average size of a tree $\overline{s}$ is
\begin{eqnarray}
\nonumber \overline{s} &=& \overline{s}\left(h=h_{\max}\right)p\left(h =
  h_{\max}\right) + \overline{s}\left(h<h_{\max}\right)p\left(h <
  h_{\max}\right),\\
\nonumber &=& \left[2^{h_{\max}+1}-1\right](1-q)^{h_{\max}-1} +\\
\nonumber && \left[\left({4q \over 2q-1}\right)\left[{1 -
  \left(2(1-q)\right)^{h_{\max}-1} \over 1 -
  (1-q)^{h_{\max}-1}}\right] - 1\right]\left[1 -
  (1-q)^{h_{\max}-1}\right],\\
&=& {2q - 2\cdot\left[2(1-q)\right]^{h_{\max}} + 1 \over 2q-1}
\end{eqnarray}
\end{enumerate}

The average size of a tree created using ramped full, ramped not-full
or ramped half-half can now be easily calculated. I have done this but
don't have time to write out the derivation here!  (I feel a bit like
Fermat ;0)

Hence, given a $q$, $maxHeight$ and GP tree initialization algorithm,
using the equations about, we can derive an estimate of average GP tree
size, $\hat{s}$. 
\end{document}